\documentclass{article}

\usepackage[main, final]{neurips_2026}

\usepackage[utf8]{inputenc} 
\usepackage[T1]{fontenc}    
\usepackage{hyperref}       
\usepackage{enumitem}
\usepackage{url}            
\usepackage{booktabs}       
\usepackage{amsfonts}       
\usepackage{nicefrac}       
\usepackage{microtype}      
\usepackage[table]{xcolor}         
\usepackage{xspace}
\usepackage{multirow}
\usepackage{wrapfig}
\usepackage{graphicx}
\usepackage{makecell}
\usepackage{subcaption}

\definecolor{lightgreen}{RGB}{200,255,200}
\definecolor{lightred}{RGB}{255,200,200}
\definecolor{lightgray}{RGB}{230,230,230}
\definecolor{myblack}{RGB}{0,0,0}
\definecolor{mydarkgreen}{HTML}{6a994e}
\definecolor{mygreen}{HTML}{6a994e}
\definecolor{myred}{HTML}{d62828}
\definecolor{mygray}{HTML}{6c757d}
\definecolor{myblue}{HTML}{4594c1}
\definecolor{mysuper}{RGB}{59,125,35}
\definecolor{mycom}{RGB}{11,119,160}
\definecolor{myinfer}{RGB}{192,79,21}
\usepackage{amsmath}
\usepackage{amssymb}
\usepackage{mathtools}
\usepackage{amsthm}
\usepackage{bm}
\usepackage[most]{tcolorbox}
\usepackage{mdframed}
\newmdenv[
  backgroundcolor=yellow!20,
  linecolor=black,
  linewidth=1pt,
]{myframe}
\newcommand{\D}{\mathcal{D}}
\newcommand{\M}{\mathcal{M}}
\newcommand{\defeq}{\vcentcolon=}
\newcommand{\Superior}{{\textcolor{mysuper}{\textbf{Superior}}}\xspace}
\newcommand{\Comparable}{{\textcolor{mycom}{\textbf{Comparable}}}\xspace}
\newcommand{\Inferior}{{\textcolor{myinfer}{\textbf{Inferior}}}\xspace}

\newcommand{\Takeaway}[1]{%
\begin{tcolorbox}[
    enhanced,
    colback=white,
    colframe=white,
    leftrule=0.4mm,
    rightrule=0.4mm,
    toprule=0.4mm,
    bottomrule=0.4mm,
    arc=0mm,
    left=0pt,
    right=0pt,
    top=2pt,
    bottom=2pt,
    breakable,
    borderline north={0.4mm}{0pt}{black},
    borderline south={0.4mm}{0pt}{black}
]
{\textbf{
{Takeaway:}}} #1
\end{tcolorbox}
}
\title{The Three Regimes of Offline-to-Online Reinforcement Learning}

\author{%
  Lu Li\textsuperscript{1,2}\quad
  Tianwei Ni\textsuperscript{1,2}\quad
  Yihao Sun\textsuperscript{1,2}\quad
  Pierre-Luc Bacon\textsuperscript{1,2,3}\quad
  \\[0.5em]
\textsuperscript{1}Mila -- Qu\'ebec AI Institute\quad
\textsuperscript{2}Universit\'e de Montr\'eal\quad 
\textsuperscript{3}CIFAR AI Chair\quad \\[0.5em]
\texttt{lu.li@mila.quebec},
\texttt{twni2016@gmail.com}
}

\begin{document}

\maketitle

\begin{abstract}
Offline-to-online reinforcement learning (RL) has emerged as a practical paradigm that leverages offline datasets for pretraining and online interactions for fine-tuning. However, its empirical behavior is highly inconsistent: design choices of online fine-tuning that work well in one setting can fail completely in another. Guided by the stability--plasticity principle, we propose a framework that can explain this inconsistency: We argue that efficient fine-tuning must preserve the utility of the stronger offline prior, whether that is the pretrained policy or the offline dataset, while maintaining sufficient plasticity. This perspective identifies three regimes of online fine-tuning, each requiring distinct stability properties. We validate this framework through a large-scale empirical study, finding that the results strongly align with its predictions in 45 out of 63 cases, with only 3 opposite mismatches. This work provides a framework for guiding design choices in offline-to-online RL based on the relative performance of the offline dataset and the pretrained policy.\looseness=-1
\end{abstract}

\section{Introduction}
Reinforcement learning (RL) has achieved impressive successes in a variety of domains~\citep{mnih2015human,AlphaGo_Zero,degrave2022magnetic}, but its reliance on large amounts of online interaction often makes direct application to real-world problems challenging. To address this challenge, recent research has turned to leveraging pre-collected datasets through offline RL or imitation learning~\citep{levine2020offline,osa2018algorithmic}, providing a strong initial policy trained from offline data. However, policies trained purely offline are often suboptimal and fail to generalize to states outside the dataset’s support, making online fine-tuning essential. Offline-to-online RL~\citep{nair2020awac, lee2022offline} addresses this issue by first pretraining an agent on an offline dataset and then fine-tuning it with additional online interactions to further improve performance.

\begin{wrapfigure}[14]{r}{0.45\linewidth}
\vspace{-1.0em}
    \centering
\includegraphics[width=\linewidth]{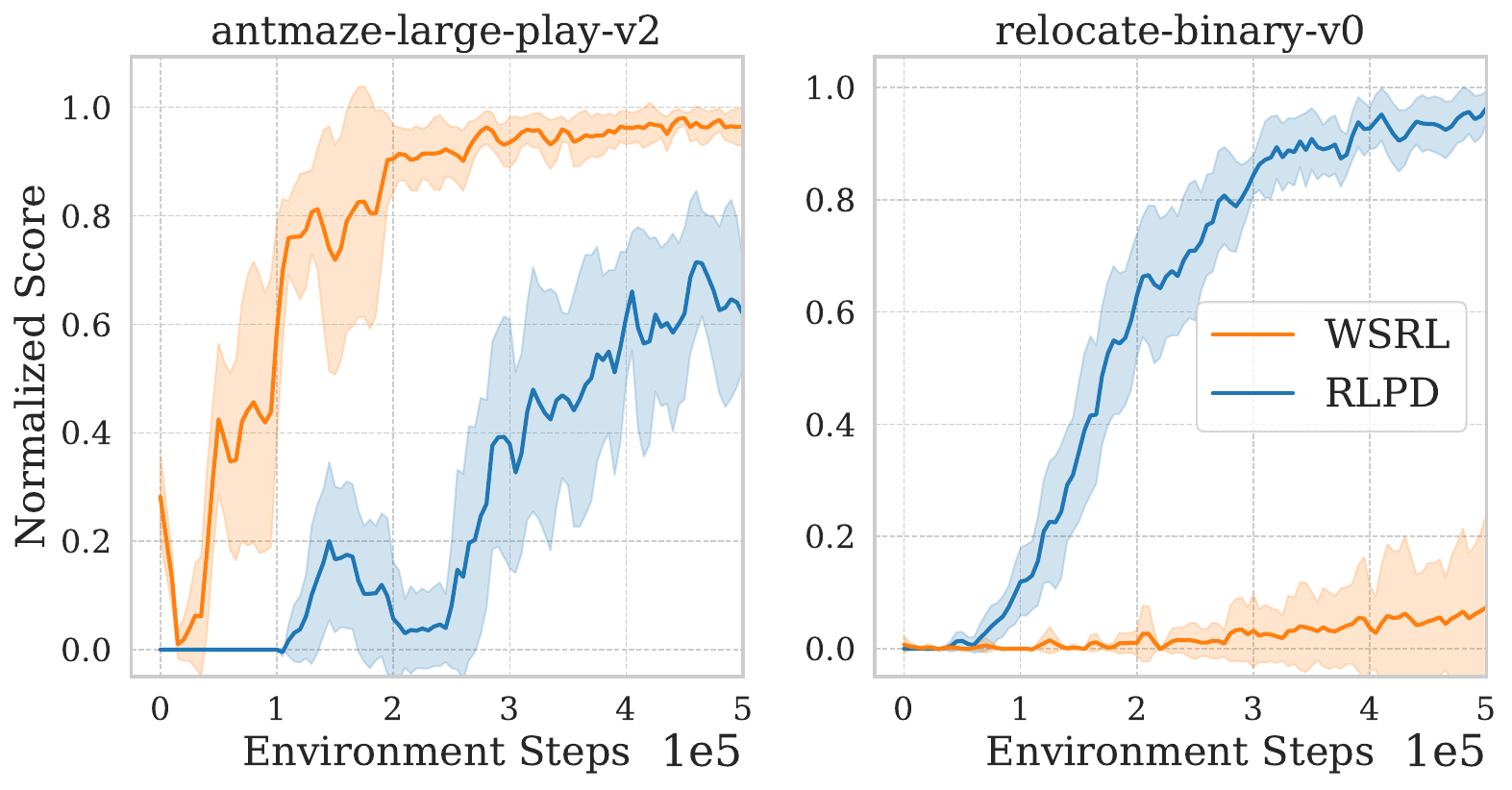}
\vspace{-1.5em}
    \caption{Comparison between \textbf{WSRL} (pretrained policy only) and \textbf{RLPD} (offline dataset only) on two representative offline-to-online RL tasks.  Learning curves are shown as mean~$\pm$~95\% CI.}
    \label{fig:wsrl_vs_rlpd}
\end{wrapfigure}

While offline-to-online RL has led to promising results, online RL fine-tuning suffers from highly inconsistent empirical behavior: design choices that work well in one setting can fail completely in another. 
For example, as shown in Figure~\ref{fig:wsrl_vs_rlpd}, on D4RL tasks~\citep{fu2020d4rl} such as \textit{antmaze-large-play-v2}, Warm-Start RL (WSRL)~\citep{zhou2024efficient}, which relies on the pretrained policy and discards the offline dataset during online fine-tuning, substantially outperforms RL with Prior Data (RLPD)~\citep{ball2023efficient}, which uses the offline dataset only at the online RL stage. In contrast, on D4RL tasks such as \textit{relocate-binary-v0}, the opposite pattern emerges, with RLPD outperforming WSRL by a wide margin. These seemingly inconsistent outcomes raise one fundamental question:
\emph{What underlying factors cause design choices to succeed in some settings but fail in others?}\looseness=-1

\begin{figure}
    \centering
    \vspace{-4.0em}
\includegraphics[width=\linewidth]{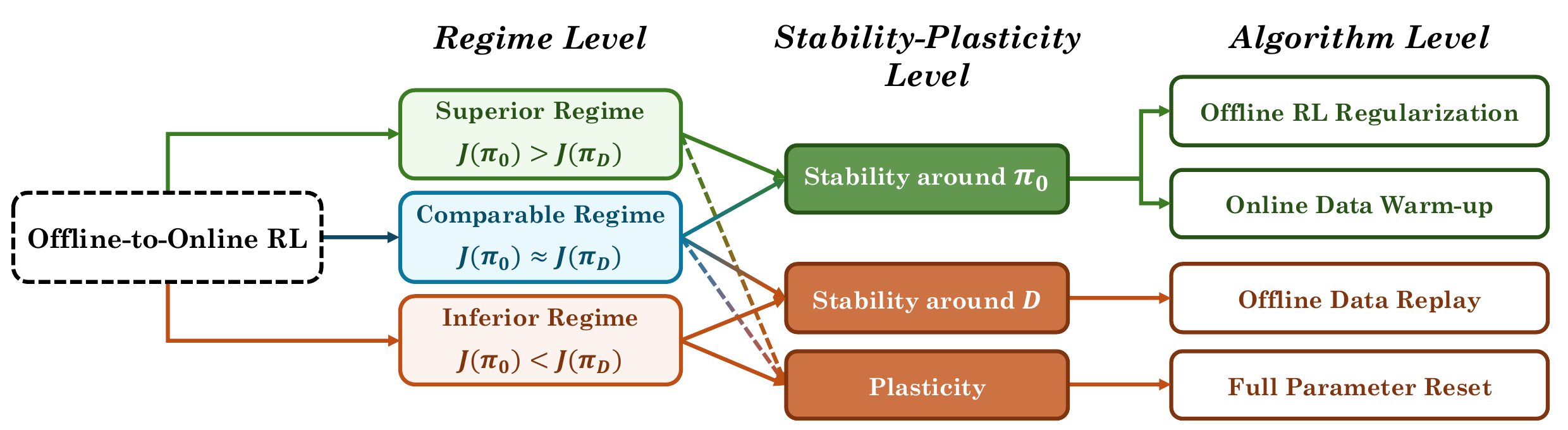}
\vspace{-1.7em}
    \caption{\textbf{Overview of the three regimes in offline-to-online RL}, defined based on the \textit{relative} performance of the pretrained policy $J(\pi_0)$ and the offline dataset $J(\pi_\mathcal{D})$. For each regime, our framework indicates which property is most needed during fine-tuning. The boxes at the right show representative design choices that implement these enhancing stability or plasticity. Dashed arrows denote weaker connections than solid arrows.}
    \vspace{-2.2em}
    \label{fig:overview}
\end{figure}

To answer this question, we propose a framework for offline-to-online RL that explains these seemingly inconsistent outcomes through the lens of the stability--plasticity principle, a perspective that has been widely studied in neuroscience~\citep{mcclelland1995there} and machine learning~\citep{kirkpatrick2017overcoming,wolczyk2024fine,dohare2024loss}.
Guided by the principle, effective fine-tuning requires a careful balance between stability and plasticity. \textbf{Stability} refers to the preservation of useful prior knowledge, ensuring that competencies acquired during pretraining are not substantially degraded. \textbf{Plasticity}, in contrast, denotes the capacity of the model to adapt flexibly and efficiently to new data. Furthermore, we identify two distinct forms of stability in offline-to-online RL: stability around the \textit{pretrained policy} $\pi_{0}$, which emphasizes preserving knowledge explicitly encoded in the policy parameters, and stability around the \textit{offline dataset} $\mathcal{D}$, which emphasizes retaining knowledge implicitly encoded in offline data. As stability and plasticity are inherently in trade-off, this distinction indicates that fine-tuning is more efficient when stability is enhanced with respect to the stronger source of offline priors, whether it is the pretrained policy or the offline dataset.\looseness-1

Building on this insight, we propose a taxonomy of \textbf{three regimes for offline-to-online RL}, each capturing a distinct relationship between the pretrained policy and the offline dataset. As shown in Figure~\ref{fig:overview}, the three regimes are defined based on which source of prior knowledge is stronger, either the pretrained policy or the offline dataset. 
Moreover, different fine-tuning methods can be systematically categorized according to whether they enhance stability around $\pi_0$, stability around $\mathcal{D}$, or plasticity. By first determining the regime, one can select or design fine-tuning strategies that match its stability-plasticity requirements. 
This yields two \textit{practical} benefits: it helps choose the most suitable method for each setting rather than applying a single uniform state-of-the-art algorithm, and it narrows the search space by indicating whether one should focus on leveraging the pretrained policy or on exploiting the offline dataset, thereby reducing unnecessary trial-and-error.

To validate this framework, we conduct a large-scale empirical study that covers 21 dataset-task compositions across four D4RL domains (MuJoCo locomotion, AntMaze navigation, Adroit manipulation, and Kitchen manipulation) and three representative pretraining algorithms, yielding 63 settings. The results align closely with the predictions of our framework, supporting its utility for guiding design choices of fine-tuning in offline-to-online RL.\looseness-1

Contributions of this paper can be summarized as:
\begin{itemize}
[leftmargin=*,itemsep=1.5pt, topsep=0pt, parsep=0pt, partopsep=0pt]
    \item We develop a diagnostic and predictive framework for offline-to-online RL guided by the stability--plasticity principle. This yields a three-regime taxonomy defined by the relative performance of the initial pretrained policy $\pi_0$, and the offline dataset $\mathcal{D}$. We also cast prior disparate fine-tuning algorithms into a unified view based on whether they enhance stability or plasticity.
    \item To validate this framework, We conduct an extensive empirical study across 63 settings. The results align with our predictions in 45 cases, with only 3 opposite mismatches. To ground these behavioral outcomes, we provide a mechanistic analysis of value learning dynamics during fine-tuning, showing that failures stem from exploding offline TD error and divergent Q-values.
    Finally, we demonstrate that our raw-return-based taxonomy consistently outperforms alternative taxonomies based on complex, indirect metrics, establishing it as a robust and practical first-order criterion.\looseness=-1
\end{itemize}

\vspace{-0.7em}
\section{Preliminary: Offline-to-Online RL}
\vspace{-0.7em}
Consider an MDP $\M = (\mathcal{S}, \mathcal{A}, P, R, \gamma)$, where the performance of a policy $\pi: \mathcal{S} \to \Delta(\mathcal{A})$ is measured by its expected discounted return: $J(\pi) = \mathbb{E}_{\pi, \M}[\sum_t \gamma^t r_t]$.  
Offline-to-online RL begins by pretraining the agent on an offline dataset $\mathcal{D}$, which is collected from $\mathcal{M}$ under an unknown behavior policy (or mixture of policies), using an offline RL algorithm $A_{\text{off}}$. This yields an offline pretrained agent whose policy is given by $\pi_{0} = A_{\text{off}}(\D)$. 
The fine-tuning step consists of using an online algorithm $A_{\text{on}}$ that starts from $\pi_0$ and interacts with $\M$ to obtain a final policy $\pi_N$ after $N$ iterations:\looseness=-1 
{\setlength{\abovedisplayskip}{0.3em}
\setlength{\belowdisplayskip}{0.3em}
\begin{equation}
\pi_N = A_{\text{on}}\left(\M, \D,\pi_{0}\right).
\end{equation}%
}

Crucially, we analyze this transition from offline pretraining to online fine-tuning as a fundamental continual learning problem~\citep{wolczyk2024fine}. Even though the underlying MDP remains identical across both phases, the agent faces a severe distributional shift from a static, external offline dataset to actively collected online experiences. This shift forces the agent to balance acquiring new behaviors against the risk of catastrophically forgetting the valuable prior knowledge extracted during pretraining~\citep{wolczyk2024fine,nakamoto2023cal}.

Although the ideal objective of offline-to-online RL is to co-design $(A_{\text{off}}, A_{\text{on}})$ to maximize $J(\pi_N)$, in this work, we focus on understanding and improving the \textbf{online RL fine-tuning} component $A_{\text{on}}$ by fixing the offline pretraining component $A_{\text{off}}$.

\vspace{-0.8em}
\section{A Decomposition of Performance Based on Stability–Plasticity Principle}
\vspace{-0.8em}
\label{section:3}

This section introduces an analytical framework for reasoning about fine-tuning in offline-to-online RL. Our goal is to characterize when and how online training leads to improvements over the offline initialization or degradations of what was already learned. We define two complementary properties of online fine-tuning: \emph{stability}, the ability to preserve previously acquired performance, and \emph{plasticity}, the capacity to improve further. 
These are grounded in the notion of a \emph{performance level}, understood as the expected return encoded either in the dataset or in the pretrained policy. 
We show that performance admits a decomposition into three terms --- offline prior, stability, and plasticity. This perspective provides both diagnostic insight and practical guidance.
\looseness=-1

\vspace{-0.6em}
\subsection{Offline Priors for Online Fine-Tuning}
\vspace{-0.6em}
We distinguish two offline priors available before online fine-tuning: the offline dataset and the pretrained policy, obtained by running an offline RL algorithm on the dataset. While the ultimate utility of these priors depends on complex factors such as state-action coverage, we abstract this complexity by evaluating both through their empirical performance. Relying on this single, first-order metric ensures \textit{practical simplicity}, providing an objective and readily quantifiable basis to directly compare two fundamentally different sources.
In Section~\ref{sec:alternative}, we compare this simple metric with alternative regime taxonomies, including dense reward proxies.\looseness=-1

\textbf{Performance of the dataset: $J(\pi_{\D})$.} 
Let $\mathcal{D}$ be the offline dataset and $\pi_{\mathcal{D}}$ be an abstract behavior policy representing the data-generating process. While $\mathcal{D}$ may be collected from a mixture of policies, $\pi_{\mathcal{D}}$ serves as a convenient abstraction. Its performance can be estimated by the average return:\looseness=-1  
{\setlength{\abovedisplayskip}{0.3em}
\setlength{\belowdisplayskip}{0.3em}
\[
J(\pi_{\mathcal{D}}) \approx \frac{1}{L} \sum_{k=1}^L \sum_{t=1}^T r_{k,t},
\]  
}where $L$ is the number of trajectories in $\D$, and $r_{k,t}$ is the reward at time step $t$ in the $k$-th trajectory. 
This measure provides a scalar summary of the return encoded in the dataset.\looseness=-1 

\textbf{Performance of the pretrained policy: $J(\pi_0)$.}  
The performance $J(\pi_0)$ reflects the inductive biases of the offline RL algorithm $A_{\text{off}}$, as well as the data quality and the underlying complexity of the MDP $\mathcal{M}$. This policy can provide a strong initialization for online fine-tuning, though it does not necessarily dominate the dataset baseline in practice. We therefore consider both $J(\pi_{\D})$ and $J(\pi_0)$ jointly as candidate sources of offline priors.\looseness=-1

\vspace{-0.6em}
\subsection{Performance Decomposition and Three Regimes}
\vspace{-0.6em}
Online fine-tuning produces a sequence of policies $\{\pi_n\}_{n=0}^N$ with corresponding performances $\{J(\pi_n)\}_{n=0}^N$. Our goal is to understand how these trajectories of performance can be expressed in terms of the offline priors identified above and the two complementary properties of stability and plasticity. This leads to a decomposition of final performance that makes explicit what is preserved from offline training and what is gained during online interaction.

\textbf{Stability with regard to a performance level.} In continual learning, stability is conventionally evaluated through an agent's ability to avoid catastrophic degradation of previously acquired capabilities~\citep{kirkpatrick2017overcoming,lopez2017gradient}. Motivated by this established standard, We define the stability of an online RL training process with respect to a performance level $l$, as the ability to retain the relative performance: 
{\setlength{\abovedisplayskip}{0.3em}
\setlength{\belowdisplayskip}{0.3em}
\begin{equation}
\text{Stability}(l) \defeq  \min\left (\min_{0\le n\le N} J(\pi_n) - l, 0\right).
\end{equation}
}This captures the worst-case performance drop during fine-tuning relative to~$l$. 
A score of zero means no degradation; a negative score measures how much was lost.

In our setting, the appropriate reference level is the best performance available from the offline pretraining phase, either from the dataset or from the pretrained policy:
{\setlength{\abovedisplayskip}{0.0em}
\setlength{\belowdisplayskip}{0.0em}
\begin{equation}
J^*_{\text{off}} \defeq \max \left( J(\pi_0),\; J(\pi_{\mathcal{D}}) \right).
\end{equation}
}%
We refer to this as the \emph{offline performance baseline}, and the stability with respect to it is:
{\setlength{\abovedisplayskip}{0.1em}
\setlength{\belowdisplayskip}{0.1em}
\begin{equation}
\text{Stability}(J^*_{\text{off}}) = \min_{0 \le n \le N} J(\pi_n) - J^*_{\text{off}} \le 0.
\end{equation}
}

\textbf{Plasticity.} In continual learning, plasticity is characterized by an agent's ability to continually adapt and improve its performance~\citep{nikishin2022primacy,dohare2024loss}. Consistent with this established notion,
we define the plasticity as the ability to improve during online fine-tuning: 
{\setlength{\abovedisplayskip}{0.3em}
\setlength{\belowdisplayskip}{0.3em}
\begin{equation}
\text{Plasticity} \defeq \max_{0\le i\le N} J(\pi_i) - \min_{0\le j\le N} J(\pi_j) \ge 0.
\end{equation}
}%
We quantify it by the largest performance gain attained, namely the gap between the best and worst observed performance. 

By relating these concepts through a \textbf{performance decomposition}, we have:\looseness=-1
\begin{myframe}
{\setlength{\abovedisplayskip}{0.3em}
\setlength{\belowdisplayskip}{0.3em}
\begin{equation*}
\label{eq:decomposition}
\underbrace{\max_{0\le n\le N} J(\pi_n)}_{\text{Best Performance}} = \underbrace{J^*_{\text{off}}}_{\text{(1)  Prior}} + 
\underbrace{\text{Stability}(J^*_{\text{off}})}_{\text{(2) Degradation} \le 0} + \underbrace{\text{Plasticity}}_{\text{(3) Online Improvement} \ge 0}.
\end{equation*}
}
\end{myframe}

This equation states that the best performance an agent achieves is the outcome of three interacting components. (1) The first term is the baseline performance established by the offline phase, either through the offline dataset or the pretrained policy, whichever is stronger. (2) The second term measures stability, which records whether this baseline is preserved or degraded during fine-tuning; it is always non-positive since performance can at best be maintained but not exceeded by this term. (3) The third term captures plasticity, defined as the performance improvement resulting from online interaction, and is non-negative by definition. Therefore, the \textit{improvement} over the prior is given by the sum of plasticity and stability.\looseness=-1


\textbf{The three regimes of offline-to-online RL.}  Given a pretrained policy $\pi_0$ and an offline dataset $\mathcal{D}$, the objective of online fine-tuning is to improve performance by balancing between stability with respect to $\max(J(\pi_0), J(\pi_\D))$ and sufficient plasticity. Based on this perspective, we identify three regimes for the online fine-tuning phase: 
\Superior: where $J(\pi_0) > J(\pi_\D)$;
\Comparable: where $J(\pi_0) \approx J(\pi_\D)$;
\Inferior: where $J(\pi_0) < J(\pi_\D)$.
These regimes are intended to reflect substantial differences in $J(\pi_0)$ and $J(\pi_\D)$, since small performance gaps may not be meaningful and therefore should not determine regime assignment. 

This regime taxonomy provides a framework for reasoning about the stability–plasticity trade-off in offline-to-online RL. It clarifies which source of prior should anchor stability in a given setting, as shown in ~\autoref{fig:overview}. In the \Superior Regime, stability relative to $\pi_0$ should be prioritized, because $\pi_0$ offers greater utility than $\mathcal{D}$. In the \Inferior Regime, stability relative to $\mathcal{D}$ should be emphasized, as $\mathcal{D}$ demonstrates more usefulness than $\pi_0$. In the \Comparable Regime, both baselines provide similar utility, so preserving either lead to similar effect. At the same time, maintaining sufficient plasticity across all regimes is essential for efficient online RL fine-tuning.

\vspace{-0.8em}
\section{Design Choices in Stability and Plasticity}
\vspace{-0.8em}
\label{section:4}

Building on the stability–plasticity principle and the regime taxonomy, we analyze concrete design choices for online RL fine-tuning. By categorizing methods according to whether they promote stability around the pretrained policy $\pi_0$, stability around the offline dataset $\mathcal{D}$, or increased plasticity, our framework organizes previously disparate practices into a structured landscape.
Because many existing algorithms entangle these components, we isolate and analyze representative modules individually to clarify their distinct effects.


\textbf{Minimal baseline.}
\label{subsec:4.1}
We begin with defining a naive online RL fine-tuning baseline that serves as the reference point for introducing additional components, which is \textit{intentionally minimalist}. It applies a standard online RL algorithm (e.g., SAC~\citep{haarnoja2018soft} or TD3~\citep{fujimoto2018addressing}) initialized with an offline-pretrained agent, without any further modifications. 
This baseline anchors the analysis and makes the marginal effect of each added component interpretable.

\vspace{-0.2em}
\subsection{Stability Relative to the Offline Dataset $\D$}
\vspace{-0.2em}

Design choices in this category promote stability by reusing the offline dataset during online fine-tuning. Incorporating $\mathcal{D}$ into the online learning process helps preserve the knowledge in the offline dataset and mitigates distribution shift between offline and online data.

A common strategy for incorporating offline data during fine-tuning is to reuse the offline dataset together with newly collected online transitions.  One approach initializes the replay buffer with the entire offline dataset, after which new online experiences are appended as the agent interacts with the environment. In this case, the ratio of offline to online data is determined by the dataset size and gradually shifts toward online data as training progresses. An alternative approach maintains two separate replay buffers: one fixed buffer containing the offline dataset and another buffer for online experiences. During training, each batch is sampled from both buffers according to a specified offline data ratio \(\alpha\). For instance, CalQL~\citep{nakamoto2023cal} and RLPD~\citep{ball2023efficient} use \(\alpha = 0.5\), corresponding to a symmetric 50\% offline and 50\% online sampling ratio.

\vspace{-0.3em}
\subsection{Stability with Respect to the Pretrained Policy $\pi_0$}
\vspace{-0.3em}

Design choices in this category focus on preserving and building upon the knowledge encoded in the pretrained policy $\pi_0$. The goal is to reduce the risk of catastrophic forgetting and ensure that fine-tuning does not erase useful behaviors learned during pretraining.  

\textbf{Online data warmup.}  
Before applying gradient updates, the agent $\pi_0$ first collects a larger amount of online data ($K$ steps)~\citep{zhou2024efficient}. This strategy reduces the mismatch between the pretraining distribution and the online data distribution, lowering the chance that early updates overwrite prior knowledge. 

\textbf{Offline RL regularization.}  
Fine-tuning can also reuse the same offline RL algorithm that produced the pretrained policy, thereby inheriting its conservative regularization. This regularization penalizes state-action pairs outside the online data. Since the online data is collected by the sequence of policies from $\pi_0$ to $\pi_N$, the regularization \textit{implicitly} anchors learning around the region visited by $\pi_0$, even if $\pi_0$ is not stored during fine-tuning. When offline data $\D$ is also used with regularization, we consider it as promoting stability towards \textit{both} $\pi_0$ and $\D$; since $\pi_0$ is derived from $\D$, this setting tends to have the strongest stability.  Such regularization is widely adopted in prior work~\citep{nair2020awac,kostrikov2021offline,tarasov2023revisiting,nakamoto2023cal}.\looseness=-1

\vspace{-0.3em}
\subsection{Plasticity: Parameter Reset} 
\vspace{-0.3em}

A direct method to increase plasticity is to reset network parameters, as randomly initialized networks tend to exhibit higher plasticity than pretrained ones~\citep{nikishin2022primacy}.
In the context of offline-to-online RL, parameter reset
can be interpreted as starting from any pretrained agent $\pi_0$ and then resetting its weights to a randomly initialized agent $\pi_1$. While this approach severely degrades initial performance (i.e., $J(\pi_1) - J(\pi_0)$ is highly negative), it significantly enhance plasticity.
RLPD~\citep{ball2023efficient} directly trains an online RL agent from random initialization, which can thus be reinterpreted as pretraining followed by full parameter reset before fine-tuning.


\vspace{-0.5em}
\section{Empirical Study}
\vspace{-0.5em}
\label{section:5}
We test the validity of our three-regime framework through a large-scale empirical study, examining whether its regime-specific predictions align with observed outcomes.  To connect the design modules described above with this framework, we group algorithms by the primary source of stability they emphasize. Methods that preserve knowledge from the pretrained policy $\pi_0$ are labeled \textbf{$\bm{\pi_0}$-centric}, while those that anchor stability to the offline dataset $\mathcal{D}$ are labeled \textbf{$\bm{\mathcal{D}}$-centric}. Approaches that combine elements of both are called \textbf{mixed $\bm{\pi_0+\mathcal{D}}$ methods}. 

We study a diverse set of benchmark tasks and dataset compositions, following the experimental protocols of prior work~\citep{nakamoto2023cal,zhou2024efficient}. Specifically, we include MuJoCo locomotion, AntMaze navigation, Adroit manipulation, and Kitchen manipulation domains from D4RL~\citep{fu2020d4rl}, covering a total of 21 dataset-task compositions. All experiments are conducted with 10 random seeds to ensure statistical reliability.\looseness-1 

\textbf{Offline pretraining phase.} We employ two representative offline RL algorithms, CalQL~\citep{nakamoto2023cal} and ReBRAC~\citep{tarasov2023revisiting}, as well as behavior cloning (BC)~\citep{schaal1996learning} using a deterministic policy. To pair a behavior-cloned policy with a critic, we pretrain the critic by Fitted Q Evaluation (FQE) ~\citep{voloshin2019empirical} after BC.
Combining the 21 dataset-task compositions with these 3 pretraining algorithms yields 63 experimental settings in total. 
Each setting is defined by a specific combination of pretraining algorithm, dataset, and task. In this work, since we focus on cases where the offline dataset and the task are from the same MDP, the pretraining algorithm and dataset are sufficient to uniquely specify a setting.\looseness=-1

We use the regime classification introduced in Section~\ref{section:3} to organize our analysis and to interpret the outcomes of the fine-tuning methods. Each of the 63 experimental settings, defined by a unique combination of pretraining algorithm, dataset, and task, is assigned to one of the three regimes based on the relative performance of the pretrained policy and the offline dataset.
Specifically, we conduct $t$-tests with a margin $\delta=0.05$ to assess whether the difference between $J(\pi_0)$ and $J(\pi_\mathcal{D})$ is statistically significant. 
The margin $\delta$ is introduced for robustness, since $J(\pi_\mathcal{D})$ is approximated by the dataset average return and small gaps between $J(\pi_0)$ and $J(\pi_\mathcal{D})$ may not be meaningful. It prevents over-interpreting numerical noise in regime assignment.
The complete set of regime assignments is reported in Table~\ref{table:dataset-pi0} in the appendix.

\textbf{Online fine-tuning phase.}
To ensure consistency between offline pretraining phase and online fine-tuning phase, we fine-tune each agent using the corresponding base algorithm. Specifically, we fine-tune CalQL-pretrained agents using SAC~\citep{haarnoja2018soft} and ReBRAC-pretrained agents using TD3~\citep{fujimoto2018addressing}. For the deterministic BC pretraining, we use TD3 for fine-tuning to match its deterministic actor structure, and use ReBRAC when applying regularization.\looseness=-1 

Since evaluating every possible design and their combinations is infeasible, we have grouped them into four categories: the minimal baseline, $\pi_0$-centric methods, $\mathcal{D}$-centric methods, and mixed $\pi_0+\mathcal{D}$ methods. 
We then evaluate six representative methods spanning these four categories, which collectively capture the key design choices explored in prior offline-to-online RL literature~\citep{nakamoto2023cal,zhou2024efficient,ball2023efficient}.
\begin{itemize}[leftmargin=*,itemsep=1.5pt, topsep=0pt, parsep=0pt, partopsep=0pt]
\item \textbf{Baseline:} Fine-tuning the pretrained policy using an online RL algorithm with only online data.\looseness=-1
\item \textbf{$\bm{\pi_0}$-centric methods:} Two variants of such methods are evaluated: the baseline with (i) online data warmup ($K = 5{,}000$ steps) and (ii) offline RL regularization using the pretraining coefficient. 
\item \textbf{$\bm{\mathcal{D}}$-centric methods:} Two variants of such methods are evaluated: the baseline with (i) offline data replay, and (ii) with offline data replay and reset. Both variants use separate replay buffers with an offline data ratio of $\alpha = 0.5$.
\item \textbf{Mixed $\bm{\pi_0+\mathcal{D}}$ methods:} The baseline combined with offline data replay and offline RL regularization, a combination widely adopted in prior work~\citep{nair2020awac,kostrikov2021offline,tarasov2023revisiting,nakamoto2023cal}.
\end{itemize}

In each setting, we focus and compare the strongest $\pi_0$-centric and $\D$-centric methods to better approximate the ideal performance achievable by each stability source, while minimizing confounding from implementation details and hyperparameter tuning. In the following subsections, We first present and analyze the fine-tuning results across regimes: the \Superior regime (Sec.~\ref{subsec:superior}), the \Inferior regime (Sec.~\ref{subsec:inferior}), and the \Comparable regime (Sec.~\ref{subsec:comparable}). Finally, in Sec.~\ref{appendix:Q-function-analysis}, we examine the value learning dynamics during fine-tuning to uncover the underlying mechanisms. Additionally, Appendix~\ref{appendix:online} details the empirical values of stability and plasticity, discussing how they align with intuition.\looseness-1

\begin{table}[t]
\small
\centering
\setlength{\tabcolsep}{3pt}
\vspace{-4em}
\caption{\textbf{Confusion matrix of fine-tuning results across the three regimes.} 
Green cells: correct predictions (45/63); red cells: opposite mismatches (3/63); gray cells: adjacent mismatches (15/63). Overall, the framework achieves \textbf{71\%} correct predictions with only \textbf{5\%} opposite mismatches.
}
\vspace{-0.1em}
\label{table:confusion}
\renewcommand{\arraystretch}{1.2}
\begin{tabular}{c|c|ccc}
\toprule
\multicolumn{2}{c|}{} & \multicolumn{3}{c}{\textbf{Pretraining Regime}} \\
\cmidrule(lr){3-5}
\multicolumn{2}{c|}{} & \Superior & \Comparable & \Inferior \\
\midrule
\multirow{3}{*}{\rotatebox{90}{\textbf{Fine-tune}}}
& $\pi_0$-centric $>$ $\mathcal{D}$-centric & \cellcolor{lightgreen}\textbf{24} & \cellcolor{lightgray}2 & \cellcolor{lightred}1 \\
& $\pi_0$-centric $\approx$ $\mathcal{D}$-centric & \cellcolor{lightgray}6 & \cellcolor{lightgreen}\textbf{2} & \cellcolor{lightgray}3 \\
& $\pi_0$-centric $<$ $\mathcal{D}$-centric & \cellcolor{lightred}2 & \cellcolor{lightgray}4 & \cellcolor{lightgreen}\textbf{19} \\
\bottomrule
\end{tabular}
\vspace{-2.5em}
\end{table}
\vspace{-1.0em}


\subsection{\Superior Regime: $J(\pi_0) > J(\pi_\D)$}
\vspace{-0.5em}
\label{subsec:superior}
In this regime, the pretrained policy $\pi_{0}$ achieves substantially higher performance than the offline dataset, which is common for suboptimal or low-quality datasets. In such cases, the offline dataset offers limited additional value, and the primary concern becomes preserving stability relative to $\pi_0$.

Representative results are shown in Figure~\ref{fig:plt_superoir}, with the complete results in this regime provided in the appendix (Figure~\ref{fig:plt_superoir_all}).
We perform $t$-tests between the strongest $\pi_0$-centric and $\D$-centric methods in each setting. The results indicate  $\pi_0$-centric methods outperform $\D$-centric methods in 24 out of 32 settings (75\%), while the remaining settings mostly show no statistically significant difference. 
These statistics correspond to the \Superior column of the confusion matrix in Table~\ref{table:confusion}.
\begin{figure*}[t]
\vspace{-4em}
    \centering
\includegraphics[width=0.85\linewidth]{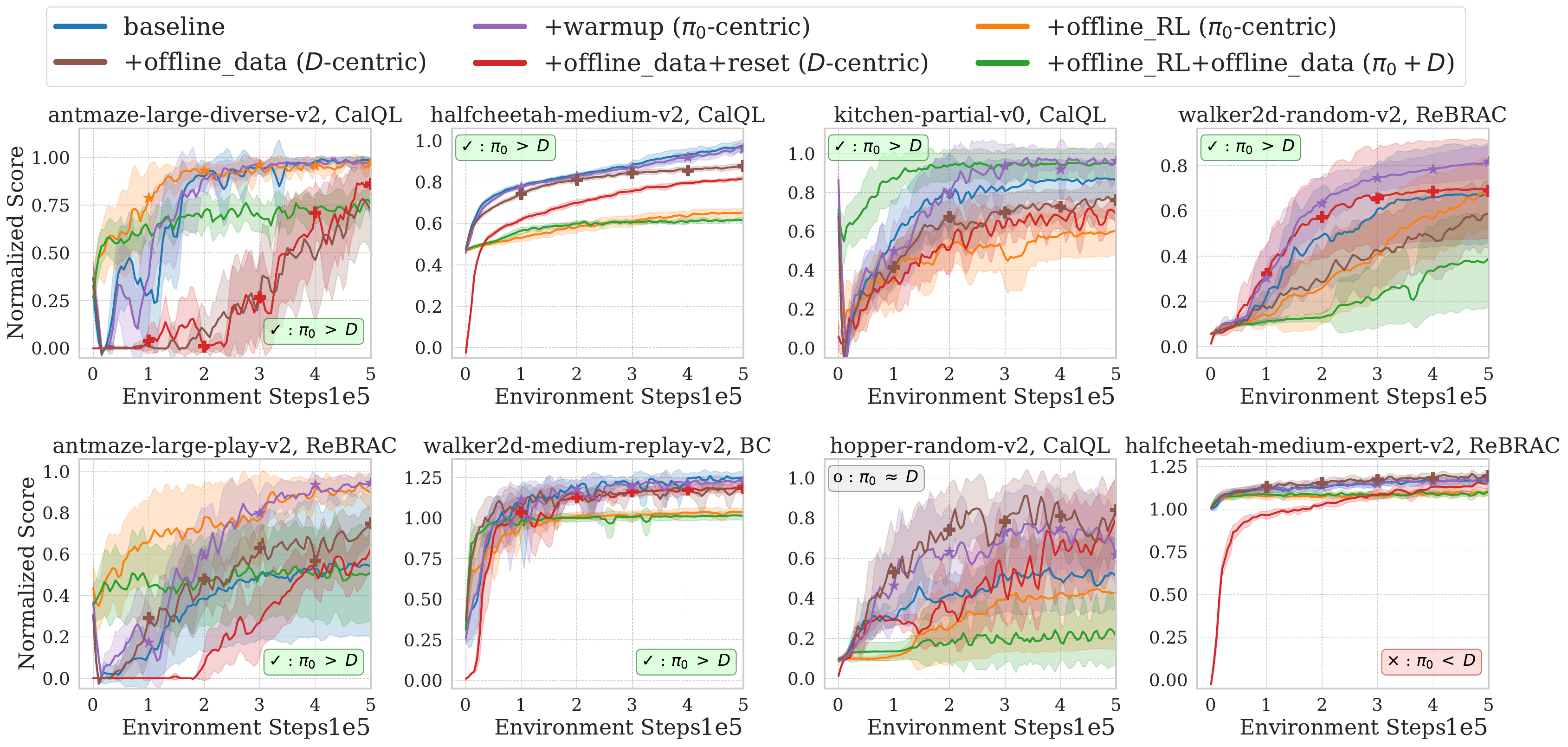}
    \vspace{-0.5em}
    \caption{Representative fine-tuning results in the \Superior regime: the first row and first two subplots in the second row are correct predictions, while the remaining two show an adjacent mismatch and an opposite mismatch. Markers on the curves indicate the better-performing variant within $\pi_{0}$-centric methods and within $\mathcal{D}$-centric methods.}
    \vspace{-1.5em}
    \label{fig:plt_superoir}
\end{figure*}

These aggregate outcomes strongly support our principle that, in the \Superior regime, $\pi_{0}$-centric methods tend to be more effective than $\mathcal{D}$-centric methods. In other words, when the pretrained policy  $\pi_0$ already outperforms the dataset, methods that stick close to $\pi_0$ work better than those that keep leaning on the offline dataset. While the prediction accuracy is not perfect, such discrepancies are anticipated given the influence of hyperparameters and implementation details. Importantly, the overall observed patterns remain consistent with our principle.\looseness-1 


Beyond aggregate comparisons, the analysis of specific design choices highlights key trade-offs between stability and plasticity.
Comparing the two $\pi_0$-centric methods, online data warmup achieves better performance in 27 out of 32 settings, reflecting its ability to preserve the pretrained policy’s knowledge while maintaining sufficient plasticity. Conversely, while offline RL regularization minimizes early performance degradation through stronger stability, it severely restricts the plasticity needed for long-term improvement. Consequently, it only outperforms warmup (5 of 32 settings) when the pretrained policy is already near-optimal. The same applies to the combination of offline RL regularization with offline data replay, which exhibits the strongest stability among all methods considered.
These contrasts highlight the importance of considering each setting and identifying the method that best balances the underlying stability–plasticity trade-off.


\Takeaway{In the \Superior regime, where the pretrained policy $\pi_0$ substantially outperforms the offline dataset $\mathcal{D}$, $\pi_0$-centric methods are typically more effective than $\mathcal{D}$-centric methods. Stronger stability proves beneficial primarily when $\pi_0$ is already close to optimal.}

\vspace{-0.5em}
\subsection{\Inferior Regime: $J(\pi_0) < J(\pi_\D)$}
\vspace{-0.5em}
\label{subsec:inferior}

In this regime, the pretrained policy $\pi_0$ performs much worse than the $\pi_\D$, which is common for near-expert datasets in sparse-reward domains. Thus, $\pi_0$ contributes substantially less useful knowledge than the offline dataset, making it crucial to retain and leverage the offline data.\looseness=-1
 
Representative results are shown in Figure~\ref{fig:plt_inferoir}, with the complete results in this regime provided in the appendix (Figure~\ref{fig:plt_inferior_all}). $\mathcal{D}$-centric methods outperform $\pi_{0}$-centric methods in 19 out of 23 settings (83\%), while the remaining settings mostly show no statistically significant difference. 
These statistics correspond to the \Inferior column of the confusion matrix in Table~\ref{table:confusion}. Taken together, these aggregate results show that in the \Inferior regime, $\mathcal{D}$-centric methods tend to be more effective than $\pi_{0}$-centric methods, consistent with the prediction of our framework.

\begin{figure*}[t]
    \centering
    \vspace{-4em}
\includegraphics[width=0.85\linewidth]{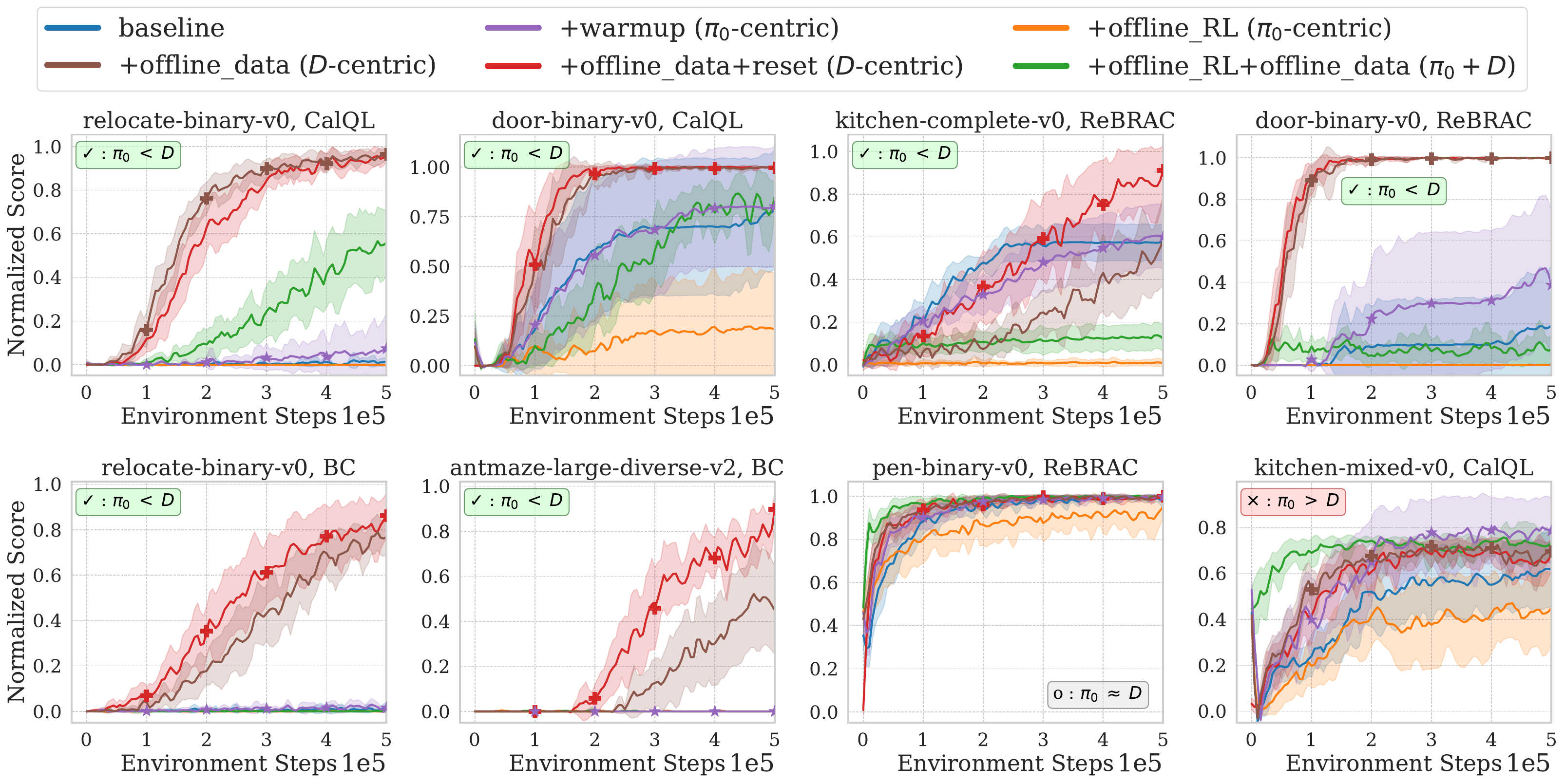}
    \vspace{-0.5em}
    \caption{Representative results in the \Inferior regime: the first six results are correct predictions, while the remaining two show an adjacent mismatch and an opposite mismatch.}
    \vspace{-2.0em}
    \label{fig:plt_inferoir}
\end{figure*}

Notably, offline data replay with reset achieves better performance than offline data replay in 13 out of 23 settings, despite the fact that reset initially causes significant degradation. This indicates that in these cases the offline pretraining phase substantially reduces plasticity while offering limited useful knowledge, and resetting the parameters allows the agent to adapt and acquire new knowledge more effectively. Furthermore, combining offline RL regularization with offline data generally underperforms compared to $\mathcal{D}$-centric methods. Although this design leverages offline data during fine-tuning, which is essential in this regime, the excessive stability limits plasticity and thereby hinders further improvement.\looseness=-1


\Takeaway{In the \Inferior regime, where the pretrained policy $\pi_0$ performs substantially worse than the offline dataset $\mathcal{D}$, $\mathcal{D}$-centric methods typically yield more effective fine-tuning. 
High plasticity (e.g., full reset) is preferred  when the pretrained agent performs  poorly.\looseness-1
}

\vspace{-0.5em}
\subsection{\Comparable  Regime: $J(\pi_0) \approx J(\pi_\D)$}
\vspace{-0.5em}
\label{subsec:comparable}
In this regime, the pretrained policy and the offline dataset yield similar performance. The complete results are provided in the appendix (Figure~\ref{fig:plt_comparable_all}).
Our framework predicts that $\pi_0$-centric and $\mathcal{D}$-centric methods should yield comparable outcomes once fully optimized. Empirically, only 2 out of 8 settings are statistically indistinguishable under $t$-tests. This seems at odds with the prediction. However, closer inspection shows that the differences are minor: in 6 of 8 settings the mean gap between categories is less than 0.1. These small gaps indicate that both anchors provide similar prior knowledge, exactly as the framework suggests.

Why, then, do mismatches arise at all? The key is that effect sizes in this regime are small by construction. When $\pi_0$ and $\mathcal{D}$ are nearly tied, outcomes become highly sensitive to hyperparameters, initialization, and other implementation details. In our study, we fixed a limited set of representative variants and hyperparameters across all settings to avoid over-tuning. This conservative design choice helps comparability but can also tip results in such close cases.

\Takeaway{In the \Comparable regime, where the pretrained policy $\pi_0$ and the offline dataset $\mathcal{D}$ exhibit similar performance, $\pi_0$-centric and $\mathcal{D}$-centric methods should in principle yield comparable fine-tuning outcomes when fully optimized, though in practice their relative performance is often sensitive to implementation details.}
\vspace{-0.5em}
\subsection{Mechanistic Analysis}
\begin{figure*}[t]
    \centering
        \vspace{-3.8em}
\includegraphics[width=0.85\linewidth]{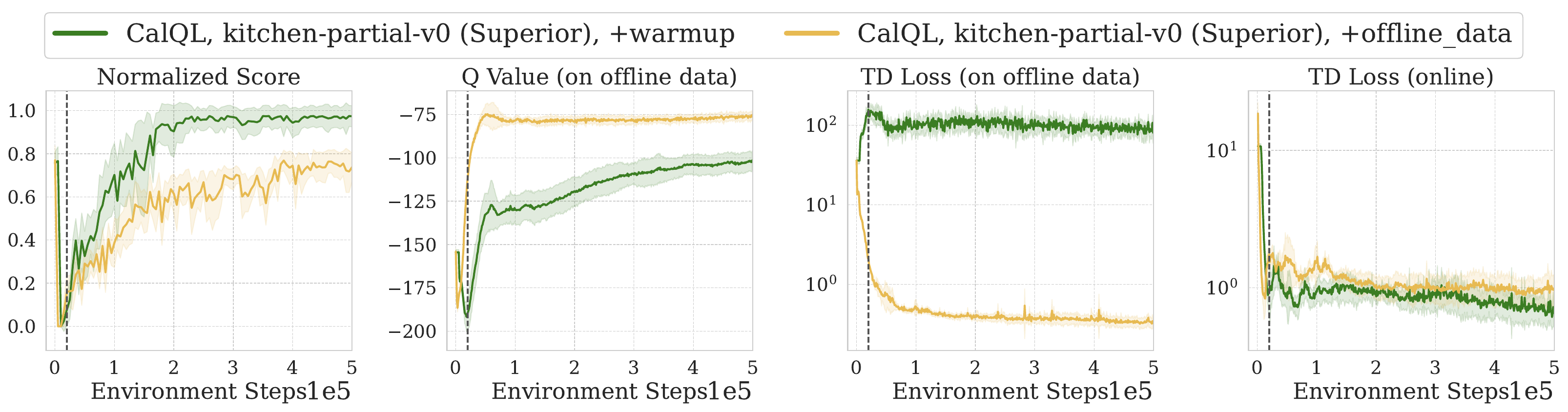}
\includegraphics[width=0.85\linewidth]{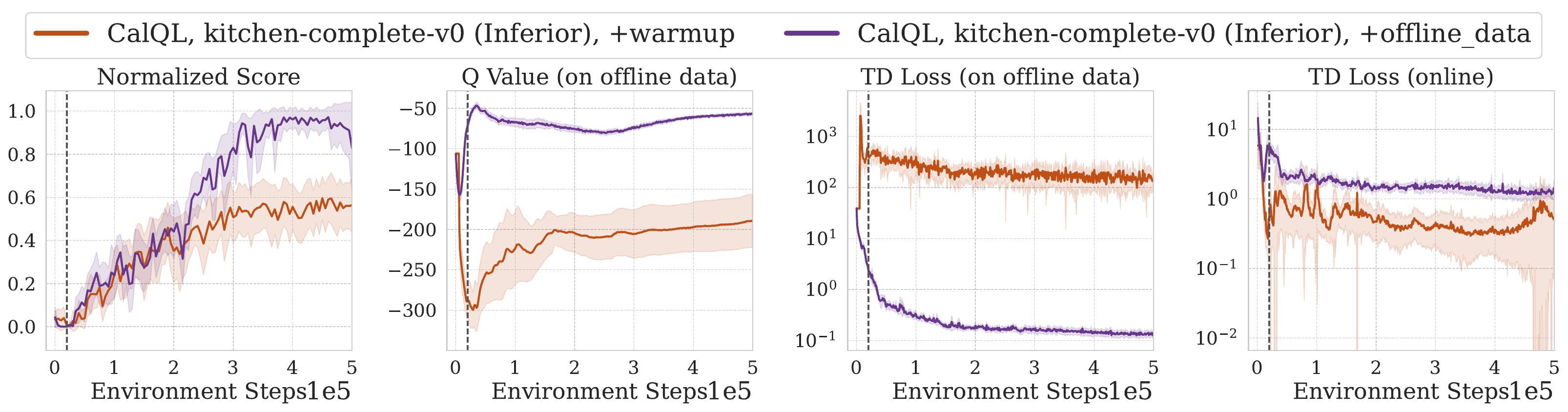}
    \vspace{-0.7em}
    \caption{Comparison of value learning dynamics in different regimes. Fine-tuning without offline data results in significantly higher TD loss on the offline dataset in the \Inferior regime compared to the \Superior regime, and also leads to divergence of the Q-values on the offline data.}
    \label{fig:online_td_q}
    \vspace{-1.5em}
\end{figure*}

\vspace{-0.5em}
\label{appendix:Q-function-analysis}
While we characterize stability and plasticity through performance metrics in our main taxonomy, it is crucial to understand the underlying mechanisms driving these behavioral outcomes. Given that value learning represents a primary bottleneck in offline-to-online RL~\citep{nakamoto2023cal,zhou2024efficient}, we analyze how the Q-function evolves during fine-tuning to uncover the mechanisms behind the \Superior and \Inferior regimes. Figure~\ref{fig:online_td_q} illustrates a representative pattern using CalQL pretraining, comparing a \Superior regime and an \Inferior regime; more results are provided in Appendix~\ref{appendix:extended-mechanistic}. In each regime, we evaluate a $\pi_0$-centric method (warmup) against a $\mathcal{D}$-centric method (offline data). To understand the mechanism about plasticity, we examine the temporal difference (TD) loss on the newly collected online data during fine-tuning. As shown in the rightmost column of Figure~\ref{fig:online_td_q}, across both regimes, fine-tuning with warmup achieves a lower online TD loss compared to using the offline dataset, mechanistically facilitating its ability to adapt to the online data distribution by imposing weaker regularization.\looseness=-1

Conversely, to understand the mechanism behind stability, we evaluate the TD loss and Q-value on data from offline dataset $\mathcal{D}$, as shown in the second and third columns of Figure 5. While fine-tuning with offline data safely anchors the critic, the warmup method lacks this anchoring and suffers a severe optimization breakdown in the \Inferior regime. Its offline TD loss exceeds $10^3$ during the first 100k steps, causing drastic Q-value divergence. Although a similar destabilization occurs in the \Superior regime, it is significantly milder, with the offline TD loss peaking near $10^2$ and avoiding Q-value divergence on the offline data.

By linking the outcomes to the value learning dynamics, this evaluation mechanistically illustrates why anchoring to the offline dataset $\mathcal{D}$ is crucial in the \Inferior regime to prevent exploding TD loss and diverging Q-values, whereas such anchoring is not necessary in the \Superior regime.



\vspace{-0.6em}
\subsection{Alternative Taxonomies of Three Regimes}
\label{sec:alternative}
\vspace{-0.6em}

To validate our raw-return metric, we benchmark against alternative regime classifications based on: (1) dense reward proxies for sparse-reward domains (Adroit, AntMaze, and Kitchen), (2) learned Q-values, and (3) BC performance. To ensure a fair comparison with raw returns, the dense rewards are constructed without access to privileged task information (e.g., maze layouts) and are used as coarse progress proxies. Full details are provided in Appendix~\ref{app:alternative_regimes}. 
As shown in Table~\ref{table:alternative-taxonomy}, our raw-return-based taxonomy outperforms the alternatives, achieving the best classification accuracy and minimizing opposite mismatches. This demonstrates that raw return provides a significantly more reliable criterion for regime identification, as it directly aligns with the true task objectives rather than relying on potentially divergent heuristic proxies.

\begingroup
\setlength{\tabcolsep}{5pt}
\renewcommand{\arraystretch}{0.9}
\begin{table}[h]

    \centering
    \small
    \vspace{-1.5em}
    \caption{Results of taxonomies based on alternative metrics beyond raw return, averaged across benchmarks. Dense-reward proxies are applied only to sparse-reward domains.}
    \vspace{-0.10em}
    \label{table:alternative-taxonomy}
    \begin{tabular}{llcc}
    \toprule
        Domains & Metric & Accuracy $\uparrow$ & Opposite mismatch $\downarrow$ \\
        \midrule
        \multirow{2}{*}{\makecell{Sparse-reward\\ (27 cases)}}
        & Raw return (ours) &  \textbf{78\%} & \textbf{4\%}\\
        & Dense-reward proxy & 65\%& 11\%\\
        \midrule
        \multirow{3}{*}{\makecell{All domains\\(63 cases)}}
        & Raw return (ours) & \textbf{71\%} & \textbf{5\%}\\
        & Q-function & 51\%&30\%\\
        & BC performance &41\% &\textbf{5\%}\\
        \bottomrule
    \end{tabular}
    \vspace{-0.5em}
\end{table}
\endgroup

\vspace{-1em}
\section{Conclusion and Discussion}
\vspace{-0.7em}
\label{section:conclusion}

This paper introduces a framework guided by the stability--plasticity principle to reconcile the puzzling variability of offline-to-online RL. We showed that the key determinant of fine-tuning success is anchoring stability to the dominant offline prior, whether that is the pretrained policy or the offline dataset. From this observation we derived a taxonomy of three regimes, each dictating where stability should be enforced and how plasticity should be managed.
The value of this framework is twofold. First, it provides a clear explanation for the conflicting empirical evidence in the literature: design choices that seem inconsistent across benchmarks in fact reflect different underlying regimes. Second, it offers actionable guidance for practitioners. By identifying the regime of a given setting, one can select methods that align with its stability–plasticity requirements, reducing reliance on trial-and-error.\looseness=-1

\textbf{Limitations and Future Work.} Our regime taxonomy provides an efficient lens for understanding offline-to-online RL by compressing complex phenomena into a small number of discrete categories. Such taxonomies have been widely used in both the natural sciences and machine learning, for example in imitation learning, where regimes have been proposed based on density ratios~\citep{spencer2021feedback} or dataset size~\citep{belkhale2023data}. 
We ground our taxonomy in raw return because task performance serves as the dominant, first-order determinant of fine-tuning dynamics in our three regimes framework. Other characteristics, such as state-action coverage of the dataset, may act as second-order factors that influence the learning process.
Extending our framework to incorporate these second-order dimensions alongside performance to yield a more comprehensive taxonomy is an important direction for future work.\looseness=-1

\section*{Acknowledgments}
Lu Li would like to thank Guozheng Ma for valuable discussions during the preparation of this work. The research was enabled in part by computational resources provided by the Digital Research Alliance of Canada (\url{https://alliancecan.ca}) and Mila (\url{https://mila.quebec}). We acknowledge funding support from CIFAR.

\small{
\bibliography{neurips_2026}
\bibliographystyle{unsrt}
}

\newpage
\appendix
\onecolumn

\makeatletter
\def\addcontentsline#1#2#3{%
  \addtocontents{a#1}{\protect\contentsline{#2}{#3}{\thepage}{\@currentHref}}%
}
\makeatother

\begingroup
  \setcounter{tocdepth}{2}
 \makeatletter
  \let\old@starttoc\@starttoc
  \def\@starttoc#1{\begingroup\parskip=2pt\old@starttoc{#1}\endgroup}
  \makeatother

  \section*{\contentsname}
  \makeatletter
  \@starttoc{atoc}
  \makeatother
\endgroup

\section{Related Work}

\subsection{Offline-to-Online RL}
Offline-to-online RL seeks to combine the strengths of offline pretraining with the adaptability of online fine-tuning.
Early approaches focused on extending offline RL regularization methods into the online regime, constraining fine-tuning updates to remain close to the pretrained policy. For example, Advantage Weighted Actor-Critic (AWAC)~\citep{nair2020awac} and Implicit Q-Learning (IQL)~\citep{kostrikov2021offline} applied offline regularization techniques directly to online fine-tuning. Building on the similar idea, PROTO~\citep{li2023proto} introduced KL regularization to explicitly constrain the online policy to the pretrained one.
Another line of work proposes new replay strategies for incorporating offline data more effectively. Lee et al.~\citep{lee2022offline} propose balanced replay to mitigate distribution shift and bootstrap error when transitioning from offline to online learning,  
EDIS\citep{liu2024energy} employs a diffusion model to select or generate samples.
Alternative strategies separate the roles of exploration and exploitation during fine-tuning. Jump-Start RL (JSRL)~\citep{uchendu2023jump}, maintains a fixed guided policy from pretraining alongside an exploration policy that is updated online, progressively transferring control from the pretrained policy to the learned one. More recent directions include expanding the action space via policy set expansion (PEX~\citep{PEX}) and Bayesian methods for uncertainty-aware exploration (BOORL~\citep{hu2024bayesian}).
Some approaches skip offline pretraining but still make use of offline data. For example, Hybrid RL~\citep{song2022hybrid} and RLPD~\citep{ball2023efficient} start training directly with online RL while incorporating offline datasets, providing another way to combine offline data with online interaction.

\subsection{Plasticity and Stability}
Plasticity and stability have long been recognized as central, often competing, objectives in learning systems. In neuroscience, this tension is formalized as the stability–plasticity dilemma~\citep{mermillod2013stability}, highlighting the challenge of integrating new knowledge without overwriting previously acquired competencies. In machine learning, similar dynamics manifest when agents must adapt to new data while preserving useful prior knowledge. Early work on continual and lifelong learning addressed this challenge via regularization techniques~\citep{kirkpatrick2017overcoming}, replay buffers~\citep{rolnick2019experience} and modular architectures~\citep{rusu2016progressive}. More recently, researchers have observed that insufficient plasticity can also hinder online deep RL, motivating methods designed to enhance plasticity of the neural network during training~\citep{nikishin2022primacy,sokar2023dormant,dohare2024loss}. 

We extend this perspective to offline-to-online RL by framing fine-tuning as a stability–plasticity trade-off between preserving knowledge from pretraining and adapting to new online data. While prior work has examined forgetting and plasticity in continual and online RL, their role in the offline-to-online transition has received limited attention. Our framework shows that stability–plasticity is not only an explanatory lens, but also yields actionable guidance by predicting which design choices are effective in different regimes.

\section{Detailed Experimental Setup and Complete Results}

\subsection{Offline Pretraining}

For offline pretraining, we train CalQL for 1M gradient steps on AntMaze, 20k on Adroit, and 250k on both Kitchen and MuJoCo locomotion tasks. ReBRAC is trained for 1M gradient steps on AntMaze, 100k on Adroit, 250k on Kitchen, and 500k on MuJoCo tasks. For the behavior cloning (BC) baseline, we perform 500K gradient steps of policy learning followed by 100k steps of fitted Q evaluation~\citep{voloshin2019empirical} (FQE) to obtain a Q-function for subsequent RL fine-tuning.
Since this work primarily focuses on the online fine-tuning stage of offline-to-online RL, we do not modify the offline pretraining algorithm or its default hyperparameters. 

For regime classification, we employ the two one-sided $t$-test (TOST) procedure with a margin of $\delta = 0.05$ and a significance level of $\alpha = 0.05$. The goal is to formally assess whether the pretrained policy $\pi_0$ and the offline dataset $\pi_{\mathcal{D}}$ are statistically indistinguishable in performance, or whether one is significantly superior. Let $\mu_0$ and $\mu_{\mathcal{D}}$ denote the mean returns of $\pi_0$ and $\pi_{\mathcal{D}}$. We conduct two one-sided tests for the null hypotheses $H_0: \mu_0 - \mu_{\mathcal{D}} \leq -\delta$ and $H_0: \mu_0 - \mu_{\mathcal{D}} \geq \delta$. If both null hypotheses are rejected, the difference is within the margin and the two are considered comparable, leading to assignment to the \Comparable Regime. If only one hypothesis is rejected, the difference is statistically significant and exceeds the margin, and the setting is assigned to either the \Superior or \Inferior Regime depending on which policy achieves the higher mean return. The statistics for each dataset and pretraining policy are reported in Table~\ref{table:dataset-pi0}.

While we use the margin parameter $\delta = 0.05$ in the main experiments, we further assess the sensitivity to   $\delta$ by comparing results under $\delta = 0$ and $\delta = 0.1$. The corresponding fine-tuning confusion matrices are reported in Table~\ref{table:confusion_delta0} and Table~\ref{table:confusion_delta0.1}.

\begin{table}[htb]
\centering
\vspace{-0em}
\caption{Confusion matrix of fine-tuning results across the three pretraining regimes with margin $\delta = 0$.}
\vspace{0.5em}
\label{table:confusion_delta0}
\small
\renewcommand{\arraystretch}{1.2}
\begin{tabular}{c|c|ccc}
\toprule
\multicolumn{2}{c|}{} & \multicolumn{3}{c}{\textbf{Pretraining Regime ($\delta=0$)}} \\
\cmidrule(lr){3-5}
\multicolumn{2}{c|}{} & \Superior & \Comparable & \Inferior \\
\midrule
\multirow{3}{*}{\rotatebox{90}{\textbf{Fine-tune}}}
& $\pi_0$-centric $>$ $\mathcal{D}$-centric & \cellcolor{lightgreen}\textbf{26} & \cellcolor{lightgray}0 & \cellcolor{lightred}1 \\
& $\pi_0$-centric $\approx$ $\mathcal{D}$-centric & \cellcolor{lightgray}8 & \cellcolor{lightgreen}\textbf{0} & \cellcolor{lightgray}3 \\
& $\pi_0$-centric $<$ $\mathcal{D}$-centric & \cellcolor{lightred}4 & \cellcolor{lightgray}1 & \cellcolor{lightgreen}\textbf{20} \\
\bottomrule
\end{tabular}
\end{table}

\begin{table}[htb]
\centering
\vspace{-1em}
\caption{Confusion matrix of fine-tuning results across the three pretraining regimes with margin $\delta = 0.1$.}
\vspace{0.5em}
\label{table:confusion_delta0.1}
\small
\renewcommand{\arraystretch}{1.2}
\begin{tabular}{c|c|ccc}
\toprule
\multicolumn{2}{c|}{} & \multicolumn{3}{c}{\textbf{Pretraining Regime ($\delta=0.1$)}} \\
\cmidrule(lr){3-5}
\multicolumn{2}{c|}{} & \Superior & \Comparable & \Inferior \\
\midrule
\multirow{3}{*}{\rotatebox{90}{\textbf{Fine-tune}}}
& $\pi_0$-centric $>$ $\mathcal{D}$-centric & \cellcolor{lightgreen}\textbf{18} & \cellcolor{lightgray}9 & \cellcolor{lightred}0 \\
& $\pi_0$-centric $\approx$ $\mathcal{D}$-centric & \cellcolor{lightgray}4 & \cellcolor{lightgreen}\textbf{5} & \cellcolor{lightgray}2 \\
& $\pi_0$-centric $<$ $\mathcal{D}$-centric & \cellcolor{lightred}2 & \cellcolor{lightgray}9 & \cellcolor{lightgreen}\textbf{14} \\
\bottomrule
\end{tabular}
\end{table}

\subsection{Online Fine-Tuning}
\label{appendix:online}
\begin{figure}[t]
    \centering
    \vspace{-0.5em}
\includegraphics[width=0.8\linewidth]{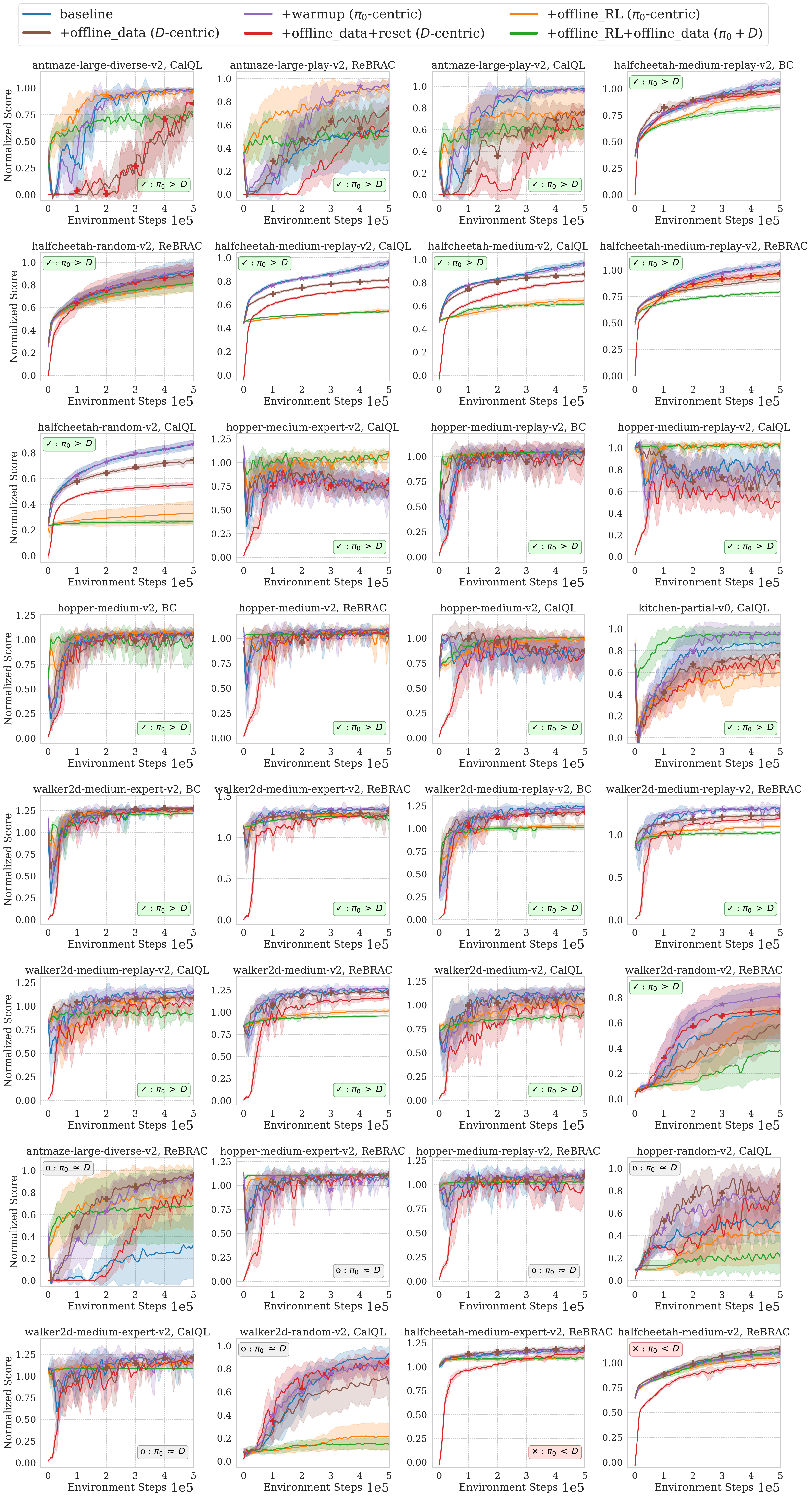}
    \caption{Full fine-tuning results in the \Superior regime.}
    \label{fig:plt_superoir_all}
\end{figure}

\begin{figure}[h]
    \centering
\includegraphics[width=0.8\linewidth]{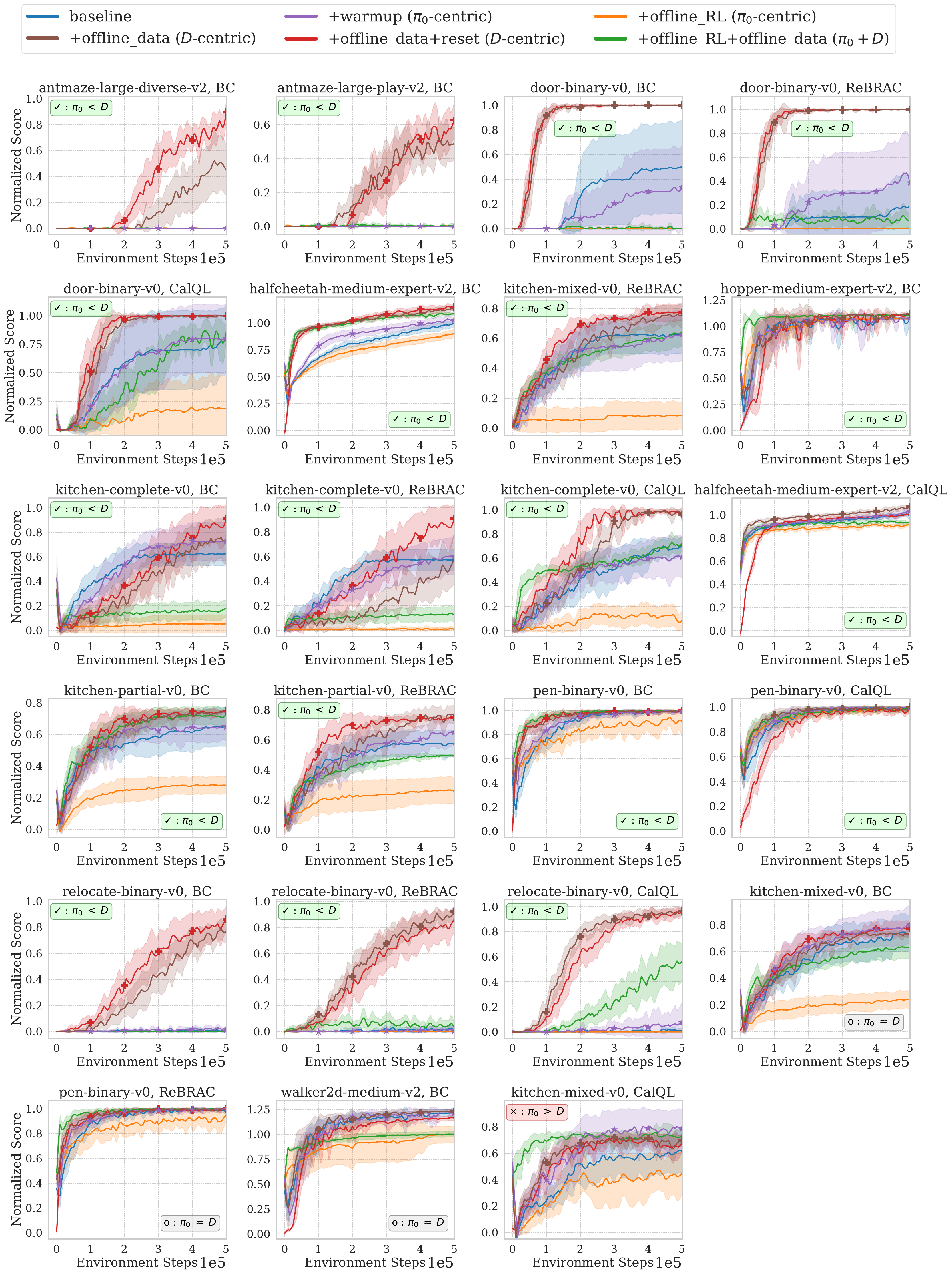}
    \caption{Full fine-tuning results in the \Inferior regime.}
    \label{fig:plt_inferior_all}
\end{figure}

\begin{figure}[h]
    \centering
\includegraphics[width=0.8\linewidth]{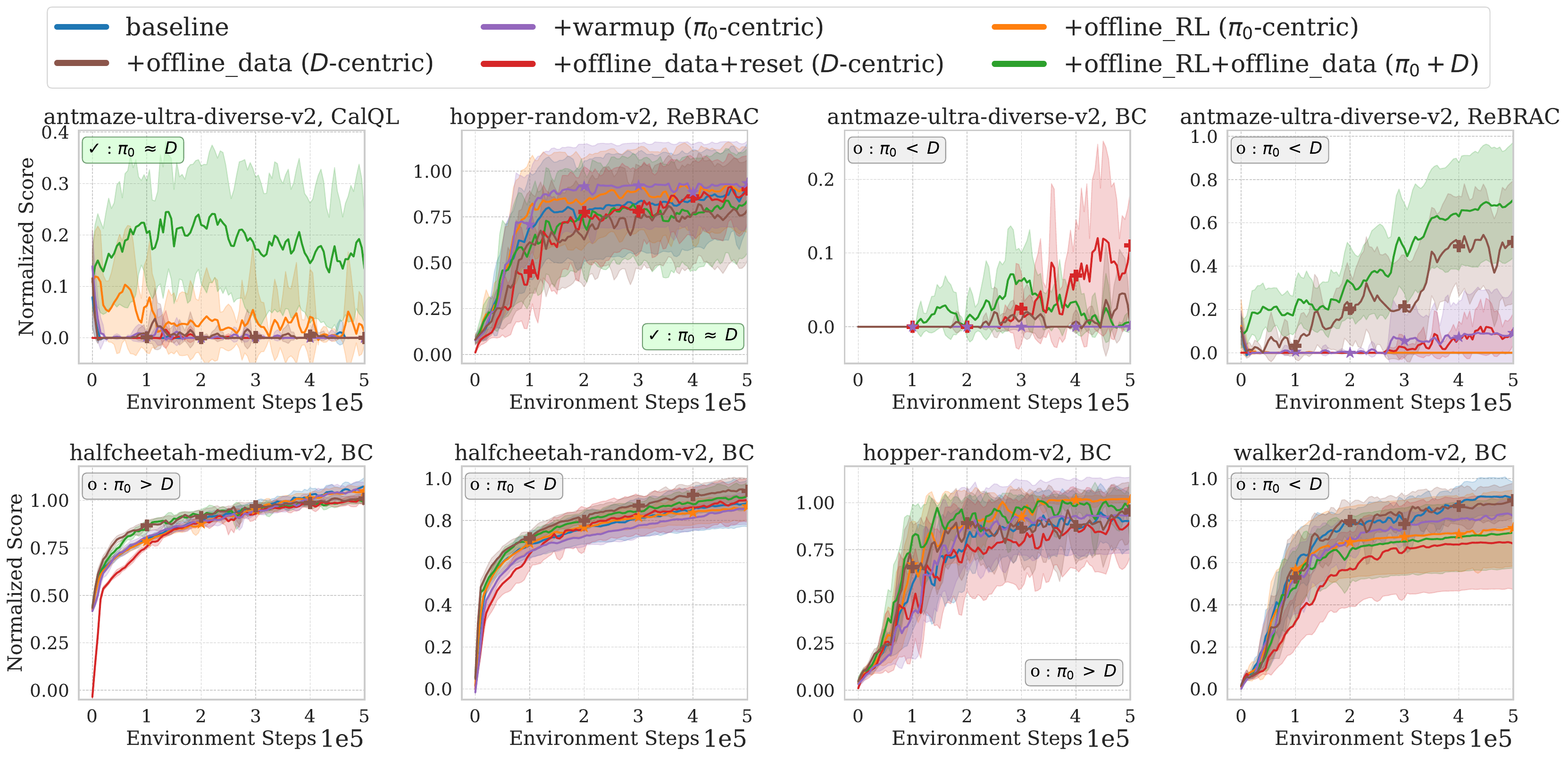}
    \caption{Full fine-tuning results in the \Comparable regime.}
    \label{fig:plt_comparable_all}
\end{figure}

Across all environments, online fine-tuning is performed for 500k environment steps with UTD=1. The complete results, categorized according to the regime taxonomy, are reported in Figure~\ref{fig:plt_superoir_all}, Figure~\ref{fig:plt_inferior_all}, and Figure~\ref{fig:plt_comparable_all}.
To obtain the strongest performance for each class in each settings, we compare the interquartile mean (IQM) of evaluation results of online data warmup and offline RL regularization within $\pi_0$-centric methods, and analogously compare offline data replay with and without reset within $\mathcal{D}$-centric methods. 
We then use the higher value from each class, comparing $\pi_0$-centric and $\mathcal{D}$-centric methods using two-sided $t$-tests with $\alpha = 0.05$.
To obtain stable and reliable $t$-test statistics, we base our analysis on the last 10 evaluation results from each random seed during online fine-tuning, which correspond to the final 50k training steps given our evaluation frequency of every 5k steps. An exception is made for \textit{door-binary-v0} and \textit{pen-binary-v0}, where we instead use results up to 200k steps, since by the end of training nearly all methods achieve a 100\% success rate, leaving no differences.

We report the empirical values of stability, plasticity, and improvement during fine-tuning for the \Superior, \Inferior, \Comparable, and all regimes. Since these quantities depend on the fine-tuning steps, we present results at both 50k and 500k environment steps, representing the \textit{early} and \textit{late} stages of fine-tuning, as shown in Table~\ref{table:superior-merged}--Table~\ref{table:merged-ordered}.

\textbf{Further discussion on early and late fine-tuning stages.}
Plasticity often correlates with performance improvement, but it is not sufficient on its own to guarantee strong results. Our analyses reveal the following regime- and stage-dependent patterns:

\begin{itemize}
[leftmargin=*,itemsep=1.5pt, topsep=0pt, parsep=0pt, partopsep=0pt]
    \item \textbf{Superior regime.} 
    The ``offline data + reset'' method attains the highest plasticity, yet yields the \emph{lowest} improvement at 50k steps and only moderate improvement at 500k steps. This shows that when $\pi_0$ is strong, plasticity alone does not determine performance; maintaining stability is essential.

    \item \textbf{Inferior regime.} 
    The relationship between stability, plasticity, and improvement depends on the fine-tuning stage. At the late stage (500k steps), methods with the highest plasticity tend to achieve the greatest improvement because $\pi_0$ performs poorly and can degrade toward near-zero performance during fine-tuning. In cases where $\min_j J(\pi_j)=0$, the improvement simplifies to $\text{plasticity} - J^*_{\text{off}}$, making plasticity the dominant factor. In contrast, at the early stage (50k steps), the ``offline RL + offline data'' method attains the highest improvement while also exhibiting the strongest stability, despite having only moderate plasticity. This indicates that stability can have a stronger influence on improvement early in fine-tuning.
\end{itemize}

\Takeaway{
Plasticity should be emphasized when $\pi_0$ is weak (\Inferior regime) and the fine-tuning budget is large (500k steps), while stability becomes more important when $\pi_0$ is strong (\Superior regime) and the fine-tuning budget is small (50k steps).
}

\begin{table}[h]
\centering
\caption{Empirical values of stability, plasticity, and improvement of different fine-tuning methods in \Superior\ Regime for 50k and 500k environment steps.}
\label{table:superior-merged}
\resizebox{0.8\textwidth}{!}{
\begin{tabular}{lccc}
\toprule
Fine-tuning method in \Superior\ regime 
& Stability $\uparrow$
& Plasticity $\uparrow$ 
& Improvement $\uparrow$ \\
\midrule
\multicolumn{4}{l}{\textbf{50k environment steps}} \\
\midrule
baseline & $-0.352 \pm 0.344$ & $0.525 \pm 0.325$ & $0.172 \pm 0.145$ \\
+ warmup ($\pi_0$-centric) & $-0.328 \pm 0.348$ & $0.501 \pm 0.315$ & $\mathbf{0.173 \pm 0.122}$ \\
+ offline RL ($\pi_0$-centric) & $-0.162 \pm 0.242$ & $0.289 \pm 0.255$ & $0.127 \pm 0.151$ \\
+ offline data ($\mathcal{D}$-centric) & $-0.293 \pm 0.351$ & $0.448 \pm 0.318$ & $0.155 \pm 0.138$ \\
+ offline data + reset ($\mathcal{D}$-centric) & $-0.615 \pm 0.338$ & $\mathbf{0.581 \pm 0.349}$ & $-0.034 \pm 0.311$ \\
+ offline RL + offline data (mixed $\pi_0+\mathcal{D}$) & $\mathbf{-0.072 \pm 0.128}$ & $0.204 \pm 0.189$ & $0.132 \pm 0.138$ \\
\midrule
\multicolumn{4}{l}{\textbf{500k environment steps}} \\
\midrule
baseline & $-0.399 \pm 0.363$ & $0.832 \pm 0.275$ & $0.433 \pm 0.276$ \\
+ warmup ($\pi_0$-centric) & $-0.394 \pm 0.371$ & $0.865 \pm 0.251$ & $\mathbf{0.471 \pm 0.258}$ \\
+ offline RL ($\pi_0$-centric) & $-0.199 \pm 0.268$ & $0.519 \pm 0.291$ & $0.319 \pm 0.256$ \\
+ offline data ($\mathcal{D}$-centric) & $-0.376 \pm 0.380$ & $0.796 \pm 0.282$ & $0.421 \pm 0.251$ \\
+ offline data + reset ($\mathcal{D}$-centric) & $-0.615 \pm 0.338$ & $\mathbf{1.011 \pm 0.183}$ & $0.396 \pm 0.271$ \\
+ offline RL + offline data (mixed $\pi_0+\mathcal{D}$) & $\mathbf{-0.124 \pm 0.213}$ & $0.393 \pm 0.288$ & $0.270 \pm 0.233$ \\
\bottomrule
\end{tabular}
}
\end{table}
\begin{table}[h]
\centering
\caption{Empirical values of stability, plasticity and improvement of different fine-tuning methods in \Inferior\ Regime for 50k and 500k environment steps.}
\label{table:inferior-merged}
\resizebox{0.8\textwidth}{!}{
\begin{tabular}{lccc}
\toprule
Fine-tuning method in \Inferior\ regime 
& Stability $\uparrow$ 
& Plasticity $\uparrow$ 
& Improvement $\uparrow$ \\
\midrule
\multicolumn{4}{l}{\textbf{50k environment steps}} \\
\midrule
baseline & $-0.688 \pm 0.348$ & $0.216 \pm 0.219$ & $-0.472 \pm 0.408$ \\
+ warmup ($\pi_0$-centric) & $-0.671 \pm 0.343$ & $0.213 \pm 0.217$ & $-0.458 \pm 0.422$ \\
+ offline RL ($\pi_0$-centric) & $-0.665 \pm 0.342$ & $0.159 \pm 0.197$ & $-0.505 \pm 0.428$ \\
+ offline data ($\mathcal{D}$-centric) & $-0.670 \pm 0.343$ & $0.268 \pm 0.215$ & $-0.402 \pm 0.398$ \\
+ offline data + reset ($\mathcal{D}$-centric) & $-0.832 \pm 0.191$ & $\mathbf{0.417 \pm 0.301}$ & $-0.415 \pm 0.395$ \\
+ offline RL + offline data (mixed $\pi_0+\mathcal{D}$) & $\mathbf{-0.640 \pm 0.357}$ & $0.266 \pm 0.219$ & $\mathbf{-0.374 \pm 0.415}$ \\
\midrule
\multicolumn{4}{l}{\textbf{500k environment steps}} \\
\midrule
baseline & $-0.692 \pm 0.307$ & $0.475 \pm 0.378$ & $-0.217 \pm 0.480$ \\
+ warmup ($\pi_0$-centric) & $-0.677 \pm 0.306$ & $0.495 \pm 0.389$ & $-0.182 \pm 0.490$ \\
+ offline RL ($\pi_0$-centric) & $-0.650 \pm 0.322$ & $0.262 \pm 0.283$ & $-0.388 \pm 0.489$ \\
+ offline data ($\mathcal{D}$-centric) & $-0.674 \pm 0.308$ & $0.696 \pm 0.340$ & $0.023 \pm 0.451$ \\
+ offline data + reset ($\mathcal{D}$-centric) & $-0.769 \pm 0.275$ & $\mathbf{0.870 \pm 0.299}$ & $\mathbf{0.101 \pm 0.205}$ \\
+ offline RL + offline data (mixed $\pi_0+\mathcal{D}$) & $\mathbf{-0.614 \pm 0.354}$ & $0.440 \pm 0.297$ & $-0.174 \pm 0.441$ \\
\bottomrule
\end{tabular}
}
\end{table}

\begin{table}[h]
\centering
\caption{Empirical values of stability, plasticity and improvement of different fine-tuning methods in \Comparable\ Regime for 50k and 500k environment steps.}
\label{table:comparable-merged}
\resizebox{0.8\textwidth}{!}{
\begin{tabular}{lccc}
\toprule
Fine-tuning method in \Comparable\ regime 
& Stability $\uparrow$ 
& Plasticity $\uparrow$ 
& Improvement $\uparrow$ \\
\midrule
\multicolumn{4}{l}{\textbf{50k environment steps}} \\
\midrule
baseline & $-0.083 \pm 0.084$ & $0.163 \pm 0.237$ & $0.080 \pm 0.251$ \\
+ warmup ($\pi_0$-centric) & $-0.082 \pm 0.085$ & $0.163 \pm 0.195$ & $0.081 \pm 0.212$ \\
+ offline RL ($\pi_0$-centric) & $-0.079 \pm 0.085$ & $0.214 \pm 0.264$ & $0.135 \pm 0.270$ \\
+ offline data ($\mathcal{D}$-centric) & $-0.081 \pm 0.085$ & $0.147 \pm 0.161$ & $0.066 \pm 0.166$ \\
+ offline data + reset ($\mathcal{D}$-centric) & $-0.114 \pm 0.147$ & $\mathbf{0.261 \pm 0.250}$ & $0.147 \pm 0.237$ \\
+ offline RL + offline data (mixed $\pi_0+\mathcal{D}$) & $\mathbf{-0.057} \pm 0.058$ & $0.240 \pm 0.251$ & $\mathbf{0.182 \pm 0.258}$ \\
\midrule
\multicolumn{4}{l}{\textbf{500k environment steps}} \\
\midrule
baseline & $-0.043 \pm 0.071$ & $0.560 \pm 0.411$ & $0.516 \pm 0.452$ \\
+ warmup ($\pi_0$-centric) & $-0.042 \pm 0.072$ & $0.543 \pm 0.390$ & $0.501 \pm 0.423$ \\
+ offline RL ($\pi_0$-centric) & $-0.043 \pm 0.072$ & $0.564 \pm 0.382$ & $0.521 \pm 0.416$ \\
+ offline data ($\mathcal{D}$-centric) & $-0.042 \pm 0.072$ & $0.680 \pm 0.338$ & $0.638 \pm 0.361$ \\
+ offline data + reset ($\mathcal{D}$-centric) & $-0.106 \pm 0.139$ & $0.623 \pm 0.431$ & $0.516 \pm 0.431$ \\
+ offline RL + offline data (mixed $\pi_0+\mathcal{D}$) & $\mathbf{-0.035 \pm 0.061}$ & $\mathbf{0.682 \pm 0.302}$ & $\mathbf{0.646 \pm 0.314}$ \\
\bottomrule
\end{tabular}
}
\end{table}

\begin{table}[h]
\centering
\caption{Empirical values of stability, plasticity and improvement of different fine-tuning methods across \textbf{all regimes} for 50k and 500k environment steps.}
\label{table:merged-ordered}
\resizebox{0.8\textwidth}{!}{
\begin{tabular}{lcccc}
\toprule
Fine-tuning method & Stability $\uparrow$ & Plasticity $\uparrow$ & Improvement $\uparrow$ \\
\midrule
\multicolumn{4}{l}{\textbf{50k environment steps}} \\
\midrule
baseline & $-0.433 \pm 0.379$ & $0.397 \pm 0.329$ & $-0.036 \pm 0.395$ \\
+ warmup ($\pi_0$-centric) & $-0.413 \pm 0.380$ & $0.382 \pm 0.315$ & $-0.031 \pm 0.389$ \\
+ offline RL ($\pi_0$-centric) & $-0.311 \pm 0.359$ & $0.242 \pm 0.246$ & $-0.069 \pm 0.404$ \\
+ offline data ($\mathcal{D}$-centric) & $-0.391 \pm 0.387$ & $0.365 \pm 0.298$ & $-0.026 \pm 0.357$ \\
+ offline data + reset ($\mathcal{D}$-centric) & $-0.632 \pm 0.349$ & $0.486 \pm 0.340$ & $-0.146 \pm 0.393$ \\
+ offline RL + offline data (mixed $\pi_0+\mathcal{D}$) & $-0.247 \pm 0.345$ & $0.226 \pm 0.206$ & $-0.021 \pm 0.357$ \\
\midrule
\multicolumn{4}{l}{\textbf{500k environment steps}} \\
\midrule
baseline & $-0.461 \pm 0.381$ & $0.667 \pm 0.375$ & $0.206 \pm 0.502$ \\
+ warmup ($\pi_0$-centric) & $-0.453 \pm 0.382$ & $0.689 \pm 0.372$ & $0.236 \pm 0.495$ \\
+ offline RL ($\pi_0$-centric) & $-0.344 \pm 0.362$ & $0.431 \pm 0.327$ & $0.087 \pm 0.526$ \\
+ offline data ($\mathcal{D}$-centric) & $-0.442 \pm 0.388$ & $0.745 \pm 0.316$ & $0.303 \pm 0.415$ \\
+ offline data + reset ($\mathcal{D}$-centric) & $-0.607 \pm 0.360$ & $0.910 \pm 0.299$ & $0.304 \pm 0.318$ \\
+ offline RL + offline data (mixed $\pi_0+\mathcal{D}$) & $-0.291 \pm 0.360$ & $0.447 \pm 0.307$ & $0.155 \pm 0.433$ \\
\bottomrule
\end{tabular}
}
\end{table}

\subsection{Hyperparameters}
We summarize the hyperparameters used in our empirical studies. The hyperparameters for SAC and CalQL are given in Table~\ref{table:sac_and_calql}. Common hyperparameters of TD3 and ReBRAC appear in Table~\ref{table:td3_and_rebrac}, while Table~\ref{table:rebrac-hp} contains the task-dependent hyperparameters for ReBRAC. Table~\ref{table:bc_and_fqe} reports the hyperparameters for BC and FQE.

\begin{table}[h]
    \caption{SAC and CalQL's hyperparameters.}
    \centering
    \label{table:sac_and_calql}
		\begin{tabular}{l|l}
			\toprule
            \textbf{Parameter} & \textbf{Value} \\
            \midrule
            optimizer                         & Adam\\
            batch size                        & 1024 \\
            learning rate       & 1e-4 \\
            Q-function soft-update rate ($\tau$)                      & 5e-3 \\
            discount factor ($\gamma$)                  & 0.999 on AntMaze, 0.99 on other \\
                \midrule
    CQL n actions & 10 \\
    CQL $\alpha$ & 1 for Adroit, 5 for others\\
    CQL max target backup & True  \\
   \bottomrule
		\end{tabular}
\end{table}

\begin{table}[h]
    \caption{Common hyperparameters of TD3 and ReBRAC.}
    \centering
    \label{table:td3_and_rebrac}
		\begin{tabular}{l|l}
			\toprule
            \textbf{Parameter} & \textbf{Value} \\
            \midrule
            optimizer                         & Adam\\
            batch size                        & 1024 \\
            learning rate       &  1e-4 on AntMaze, 1e-3 on other \\
            Q-function soft-update rate ($\tau$)                      & 5e-3 \\
            discount factor ($\gamma$)                  & 0.999 on AntMaze, 0.99 on other \\
   \bottomrule
		\end{tabular}
\end{table}

\begin{table}[h]
\caption{ReBRAC hyperparameters used in our experiments. All hyperparameters follow the best hyperparameters reported in ReBRAC~\citep{tarasov2023revisiting}, except for the Kitchen domain, which was not included; for Kitchen, we performed a hyperparameter search and selected the best configuration.}
    \label{table:rebrac-hp}
    \centering
		\begin{tabular}{l|r|r}
			\toprule
            \textbf{Task Name} & $\beta_1$ (actor) & $\beta_2$ (critic)\\%
            \midrule
            halfcheetah-random & 0.001 & 0.1 \\
            halfcheetah-medium & 0.001 & 0.01 \\
            halfcheetah-medium-expert & 0.01 & 0.1 \\
            halfcheetah-medium-replay & 0.01 & 0.001 \\
            \midrule
            hopper-random & 0.001 & 0.01 \\
            hopper-medium & 0.01 & 0.001 \\
            hopper-medium-expert & 0.1 & 0.01 \\
            hopper-medium-replay & 0.05 & 0.5 \\
            \midrule
            walker2d-random & 0.01 & 0.0 \\
            walker2d-medium & 0.05 & 0.1\\
            walker2d-medium-expert & 0.01 & 0.01 \\
            walker2d-medium-replay & 0.05 & 0.01\\
            \midrule
            antmaze-large-play     & 0.002 & 0.001 \\
            antmaze-large-diverse  & 0.002 & 0.002 \\
            antmaze-ultra-diverse  & 0.002 & 0.002 \\
            \midrule
            door-binary & 0.1 & 0.01 \\
            pen-binary & 0.1 & 0.01 \\
            relocate-binary &  0.1 & 0.01 \\
            \midrule
            kitchen-complete & 0.1& 0.001\\
            kitchen-mixed & 0.1& 0.001 \\
            kitchen-partial & 0.1& 0.001 \\
            \bottomrule
		\end{tabular}
    \vskip -0.2in
\end{table}

\begin{table}[h]
    \caption{BC and FQE's hyperparameters.}
    \centering
    \label{table:bc_and_fqe}
		\begin{tabular}{l|l}
			\toprule
            \textbf{Parameter} & \textbf{Value} \\
            \midrule
            optimizer                         & Adam\\
            batch size                        & 1024 \\
            learning rate       &  3e-4 \\
            action & deterministic \\
            FQE steps& 1e5\\
            Q-function soft-update rate ($\tau$) & 5e-3\\
            discount factor ($\gamma$) & 0.999 on AntMaze, 0.99 on other\\
   \bottomrule
		\end{tabular}
\end{table}

\subsection{Compute Details}
All experiments were conducted on a single-GPU setup using an NVIDIA L40S GPU, 24 CPU workers, and
20GB of RAM.

\begin{table}[h]
\centering
\caption{
Statistics (mean $\pm$ std) of offline datasets and pretrained policies. 
Pretrained policy scores $J(\pi_0)$ are reported as averages over 10 random seeds. 
}
\label{table:dataset-pi0}
\resizebox{\textwidth}{!}{%
\begin{tabular}{lcccccccc}
\toprule
\multirow{2}{*}{\textbf{Dataset}} &
\multicolumn{2}{c}{\textbf{Dataset}} &
\multicolumn{2}{c}{\textbf{CalQL}} &
\multicolumn{2}{c}{\textbf{ReBRAC}} &
\multicolumn{2}{c}{\textbf{BC}} \\
\cmidrule(lr){2-3}\cmidrule(lr){4-5}\cmidrule(lr){6-7}\cmidrule(lr){8-9}
& $J(\pi_\D)$ & \#Trajs & $J(\pi_0)$ & order & $J(\pi_0)$ & order & $J(\pi_0)$ & order \\
\midrule
halfcheetah-random-v2      & -0.001 $\pm$ 0.006 & 1000 & 0.248 $\pm$ 0.014 & $>$ & 0.275 $\pm$ 0.011 & $>$ & 0.019 $\pm$ 0.001 & $\approx$ \\
halfcheetah-medium-replay-v2 & 0.271 $\pm$ 0.135 & 202 & 0.451 $\pm$ 0.002 & $>$ & 0.504 $\pm$ 0.003 & $>$ & 0.370 $\pm$ 0.007 & $>$ \\
halfcheetah-medium-v2      & 0.406 $\pm$ 0.029 & 1000 & 0.470 $\pm$ 0.003 & $>$ & 0.651 $\pm$ 0.010 & $>$ & 0.427 $\pm$ 0.002 & $\approx$ \\
halfcheetah-medium-expert-v2 & 0.643 $\pm$ 0.239 & 2000 & 0.519 $\pm$ 0.078 & $<$ & 1.010 $\pm$ 0.019 & $>$ & 0.563 $\pm$ 0.025 & $<$ \\
\midrule
hopper-random-v2           & 0.012 $\pm$ 0.005 & 45240 & 0.091 $\pm$ 0.017 & $>$ & 0.082 $\pm$ 0.036 & $\approx$ & 0.038 $\pm$ 0.022 & $\approx$ \\
hopper-medium-replay-v2    & 0.150 $\pm$ 0.157 & 2039 & 1.001 $\pm$ 0.009 & $>$ & 0.965 $\pm$ 0.042 & $>$ & 0.398 $\pm$ 0.037 & $>$ \\
hopper-medium-v2           & 0.443 $\pm$ 0.117 & 2187 & 0.672 $\pm$ 0.039 & $>$ & 1.016 $\pm$ 0.012 & $>$ & 0.554 $\pm$ 0.010 & $>$ \\
hopper-medium-expert-v2    & 0.648 $\pm$ 0.319 & 3214 & 1.058 $\pm$ 0.108 & $>$ & 1.063 $\pm$ 0.042 & $>$ & 0.556 $\pm$ 0.012 & $<$ \\
\midrule
walker2d-random-v2         & 0.000 $\pm$ 0.001 & 48908 & 0.082 $\pm$ 0.043 & $>$ & 0.058 $\pm$ 0.001 & $>$ & 0.008 $\pm$ 0.001 & $\approx$ \\
walker2d-medium-replay-v2  & 0.148 $\pm$ 0.195 & 1093 & 0.843 $\pm$ 0.025 & $>$ & 0.853 $\pm$ 0.045 & $>$ & 0.276 $\pm$ 0.074 & $>$ \\
walker2d-medium-v2         & 0.620 $\pm$ 0.239 & 1191 & 0.742 $\pm$ 0.071 & $>$ & 0.845 $\pm$ 0.008 & $>$ & 0.507 $\pm$ 0.073 & $<$ \\
walker2d-medium-expert-v2  & 0.826 $\pm$ 0.285 & 2191 & 1.073 $\pm$ 0.035 & $>$ & 1.113 $\pm$ 0.004 & $>$ & 1.075 $\pm$ 0.004 & $>$ \\
\midrule
pen-binary-v0              & 1.000 $\pm$ 0.000 & 846 & 0.657 $\pm$ 0.059 & $<$ & 0.451 $\pm$ 0.086 & $<$ & 0.589 $\pm$ 0.068 & $<$ \\
door-binary-v0             & 1.000 $\pm$ 0.000 & 82 & 0.112 $\pm$ 0.123 & $<$ & 0.000 $\pm$ 0.000 & $<$ & 0.000 $\pm$ 0.000 & $<$ \\
relocate-binary-v0         & 1.000 $\pm$ 0.000 & 36 & 0.010 $\pm$ 0.011 & $<$ & 0.000 $\pm$ 0.000 & $<$ & 0.000 $\pm$ 0.000 & $<$ \\
\midrule
kitchen-partial-v0         & 0.586 $\pm$ 0.187 & 600 & 0.764 $\pm$ 0.094 & $>$ & 0.133 $\pm$ 0.085 & $<$ & 0.222 $\pm$ 0.079 & $<$ \\
kitchen-mixed-v0           & 0.598 $\pm$ 0.145 & 600 & 0.464 $\pm$ 0.092 & $<$ & 0.034 $\pm$ 0.033 & $<$ & 0.275 $\pm$ 0.049 & $<$ \\
kitchen-complete-v0        & 1.000 $\pm$ 0.000 & 19 & 0.043 $\pm$ 0.078 & $<$ & 0.002 $\pm$ 0.004 & $<$ & 0.385 $\pm$ 0.202 & $<$ \\
\midrule
antmaze-large-diverse-v2   & 0.106 $\pm$ 0.308 & 999 & 0.305 $\pm$ 0.055 & $>$ & 0.399 $\pm$ 0.080 & $>$ & 0.000 $\pm$ 0.000 & $<$ \\
antmaze-large-play-v2      & 0.105 $\pm$ 0.307 & 999 & 0.247 $\pm$ 0.068 & $>$ & 0.351 $\pm$ 0.072 & $>$ & 0.000 $\pm$ 0.000 & $<$ \\
antmaze-ultra-diverse-v2   & 0.053 $\pm$ 0.224 & 999 & 0.118 $\pm$ 0.072 & $\approx$ & 0.127 $\pm$ 0.132 & $\approx$ & 0.000 $\pm$ 0.000 & $\approx$ \\
\bottomrule
\end{tabular}
}
\end{table}

\clearpage

\clearpage

\section{Alternative Taxonomies of Three Regimes}
\label{app:alternative_regimes}
While our framework is defined using the returns of the pretrained agents and the dataset, it is also useful to consider alternative taxonomies based on different metrics. Below we present three such taxonomies, derived from dense reward proxy, Q-functions, and behavior-cloning performance, respectively.

\subsection{Dense-Reward-Based Taxonomy}

Several domains in our study (AntMaze, Adroit, and Kitchen) feature sparse rewards, where raw episode returns can be an incomplete proxy for the usefulness of prior knowledge. In particular, a pretrained policy may achieve low or zero success rate while still consistently reaching meaningful intermediate states that help exploration and fine-tuning. In such cases, classifying regimes solely based on sparse returns may underestimate the value of pretrained policy.

To address this limitation, we introduce a dense-reward-based taxonomy as a complementary regime classifier. We construct task-specific \textit{dense reward proxies} using limited environmental knowledge (e.g., distance-to-goal, object proximity, or task progress), without assuming access to oracle dense signals. These dense rewards are used only for regime identification and analysis, not for training.

\paragraph{Dense reward proxy for AntMaze.} We define a simple dense proxy as the negative Euclidean distance between the current agent position $x$ and the goal location $g$:
\begin{equation}
    r_{\text{maze}} = - \| x - g \|_2.
\end{equation}
While AntMaze contains internal walls that make Euclidean distance an imperfect measure of true task progress, we intentionally avoid using shortest-path signals. This choice mirrors the sparse-reward metric by restricting access to privileged environment information, while still providing a coarse indicator of progress toward the goal.

\paragraph{Dense reward proxy for Adroit.} We construct dense reward proxies for the Pen, Relocate, and Door tasks based on their respective success criteria. In each case, we formulate the reward as a negative squared penalty for deviations from the success thresholds.

\begin{itemize}
[leftmargin=*,itemsep=1.5pt, topsep=0pt, parsep=0pt, partopsep=0pt]
    \item Pen: Success is defined by an object-goal distance $d < 0.075$ and orientation similarity $s > 0.95$. We define the proxy as:
    \begin{equation}
        r_{\text{pen}} = - \max(d - 0.075, 0)^2 - \max(0.95 - s, 0)^2.
    \end{equation}

    \item Relocate: Success requires the object-target distance $d$ to be less than $0.1$. The proxy is defined as:
    \begin{equation}
        r_{\text{relocate}} = -\max(d - 0.1, 0)^2.
    \end{equation}

    \item Door: The goal is achieved when the door position $p$ exceeds $1.35$. We penalize positions below this threshold:
    \begin{equation}
        r_{\text{door}} = -\min(p - 1.35, 0)^2.
    \end{equation}
\end{itemize}

\paragraph{Dense reward proxy for Kitchen.} Kitchen involves a sequence of subtasks, for which we design a dense reward that incentivizes sequential completion. Let $\mathcal{T}$ be the ordered set of subtasks. For each subtask $i \in \mathcal{T}$, the reward component $r_i$ is determined by its current status:
\begin{equation}
    r_i = 
    \begin{cases} 
        0 & \text{if subtask } i \text{ is completed,} \\
        -\min\left(\frac{d_i}{d_{i,\text{init}}}, 1\right) & \text{if subtask } i \text{ is currently active,} \\
        -1 & \text{if subtask } i \text{ is a future task.}
    \end{cases}
\end{equation}
Here $d_i$ is the current Euclidean distance to the subtask goal, and $d_{i,\text{init}}$ is the distance recorded when the subtask first became active. The total dense reward is the sum over all subtasks: $r_{\text{kitchen}} = \sum_{i \in \mathcal{T}} r_i$.

\paragraph{Accuracy on sparse-reward domains.} Following our primary methodology, we employ a $t$-test to assess whether the performance $J(\pi_0)$ and $J(\pi_{\mathcal{D}})$ differs significantly under the dense reward metric. However, unlike the return-based formulation, we omit the significance margin $\delta$ since the widely varying scales of dense rewards across different domains. Based on this $t$-test, we classify each task into the three regimes. The resulting confusion matrix for the dense-reward taxonomy is presented in Table~\ref{table:dense-reward-regime-confusion}; it correctly identifies the regime in 16 out of 27 cases (59\%). For comparison, we evaluate our proposed taxonomy within the same sparse-reward settings using margins of $\delta=0.05$ and $\delta=0$ (Table~\ref{table:our-regime-confusion-sparse-05-margin} and Table~\ref{table:our-regime-confusion-sparse-no-margin}, respectively). Notably, both configurations achieve an accuracy of \textbf{78\%}, substantially outperforming the dense-reward-based alternative.


\begin{table}[htb]
\small
\centering
\caption{Confusion matrix of fine-tuning results using \textbf{dense-reward-based taxonomy in sparse-reward domains}. 
Green cells: correct predictions (16/27); red cells: opposite mismatches (3/27); gray cells: adjacent mismatches (8/27). Overall, dense-reward-based taxonomy achieves 59\% correct predictions with 11\% opposite mismatches.}
\vspace{0.5em}
\label{table:dense-reward-regime-confusion}
\renewcommand{\arraystretch}{1.2}
\begin{tabular}{c|c|ccc}
\toprule
\multicolumn{2}{c|}{} & \multicolumn{3}{c}{\textbf{dense-reward-based Regime}} \\
\cmidrule(lr){3-5}
\multicolumn{2}{c|}{} & \Superior & \Comparable & \Inferior \\
\midrule
\multirow{3}{*}{\rotatebox{90}{\textbf{Fine-tune}}}
& $\pi_0$-centric $>$ $\mathcal{D}$-centric & \cellcolor{lightgreen}\textbf{4} & \cellcolor{lightgray}0 & \cellcolor{lightred}1 \\
& $\pi_0$-centric $\approx$ $\mathcal{D}$-centric & \cellcolor{lightgray}2 & \cellcolor{lightgreen}\textbf{0} & \cellcolor{lightgray}2 \\
& $\pi_0$-centric $<$ $\mathcal{D}$-centric & \cellcolor{lightred}2 & \cellcolor{lightgray}4 & \cellcolor{lightgreen}\textbf{12} \\
\bottomrule
\end{tabular}
\end{table}

\begin{table*}[htb]
\small
\centering
\caption{Confusion matrices of fine-tuning results using \textbf{raw-return-based taxonomy (ours) in sparse-reward domains}. We provide our taxonomy using a margin of $\delta=0.05$ (Left) and $\delta=0$ (Right). Green cells indicate correct predictions, red cells denote opposite mismatches, and gray cells represent adjacent mismatches. In both settings, our taxonomy achieves \textbf{78\% accuracy} (21/27 correct) with only 4\% opposite mismatches.}
\label{table:combined-confusion-sparse}
\renewcommand{\arraystretch}{1.2}

\begin{subtable}{0.48\textwidth}
    \centering
    \caption{With Margin ($\delta=0.05$)}
    \label{table:our-regime-confusion-sparse-05-margin}
    \resizebox{\linewidth}{!}{
    \begin{tabular}{c|c|ccc}
    \toprule
    \multicolumn{2}{c|}{} & \multicolumn{3}{c}{\textbf{Pretraining Regime (Ours)}} \\
    \cmidrule(lr){3-5}
    \multicolumn{2}{c|}{} & \Superior & \Comparable & \Inferior \\
    \midrule
    \multirow{3}{*}{\rotatebox{90}{\textbf{Fine-tune}}}
    & $\pi_0$-centric $>$ $\mathcal{D}$-centric & \cellcolor{lightgreen}\textbf{4} & \cellcolor{lightgray}0 & \cellcolor{lightred}1 \\
    & $\pi_0$-centric $\approx$ $\mathcal{D}$-centric & \cellcolor{lightgray}1 & \cellcolor{lightgreen}\textbf{1} & \cellcolor{lightgray}2 \\
    & $\pi_0$-centric $<$ $\mathcal{D}$-centric & \cellcolor{lightred}0 & \cellcolor{lightgray}2 & \cellcolor{lightgreen}\textbf{16} \\
    \bottomrule
    \end{tabular}
    }
\end{subtable}
\hfill
\begin{subtable}{0.48\textwidth}
    \centering
    \caption{No Margin ($\delta=0$)}
    \label{table:our-regime-confusion-sparse-no-margin}
    \resizebox{\linewidth}{!}{
    \begin{tabular}{c|c|ccc}
    \toprule
    \multicolumn{2}{c|}{} & \multicolumn{3}{c}{\textbf{Pretraining Regime (Ours)}} \\
    \cmidrule(lr){3-5}
    \multicolumn{2}{c|}{} & \Superior & \Comparable & \Inferior \\
    \midrule
    \multirow{3}{*}{\rotatebox{90}{\textbf{Fine-tune}}}
    & $\pi_0$-centric $>$ $\mathcal{D}$-centric & \cellcolor{lightgreen}\textbf{4} & \cellcolor{lightgray}0 & \cellcolor{lightred}1 \\
    & $\pi_0$-centric $\approx$ $\mathcal{D}$-centric & \cellcolor{lightgray}2 & \cellcolor{lightgreen}\textbf{0} & \cellcolor{lightgray}2 \\
    & $\pi_0$-centric $<$ $\mathcal{D}$-centric & \cellcolor{lightred}0 & \cellcolor{lightgray}1 & \cellcolor{lightgreen}\textbf{17} \\
    \bottomrule
    \end{tabular}
    }
\end{subtable}
\end{table*}

\subsection{Q-Function-Based Taxonomy}

Conservative offline RL aims to outperform the behavior policy that generates the offline dataset~\citep{levine2020offline}. Ideally, the value function of the pretrained policy $\pi_0$ should therefore exceed that of the behavior policy $\pi_\D$ on in-dataset state-action pairs. 
Formally, Kostrikov et al.~\citep[Lemma 2]{kostrikov2021offline} show that for in-sample Q-learning, the optimal value function satisfies
\begin{equation}
\label{eq:iql_sa}
Q^{\pi_0^*}(s,a) \ge Q^{\pi_\D}(s,a), \quad \forall (s,a) \in \D,
\end{equation}
where $\pi_0^*$ denotes the optimal pretrained policy and $Q^\pi$ is the ground-truth value function of policy $\pi$. A weaker, expectation-based version of this condition is
\begin{equation}
\label{eq:iql_D}
\mathbb{E}_{(s,a) \sim \D}\!\left[Q^{\pi_0^*}(s,a)\right] \ge \mathbb{E}_{(s,a) \sim \D}\!\left[ Q^{\pi_\D}(s,a)\right] .
\end{equation}

Motivated by this result, we examine a taxonomy based on comparing $\mathbb{E}_{(s,a) \sim \D}[\hat Q^{\pi_0}(s,a)]$ and $\mathbb{E}_{(s,a) \sim \D}[Q^{\pi_\D}(s,a)]$, where $\hat Q^{\pi_0}$ denotes the pretrained Q-function.  
The intuition is that if $$\mathbb{E}_{(s,a) \sim \D}[\hat Q^{\pi_0}(s,a)] \ge \mathbb{E}_{(s,a) \sim \D}[Q^{\pi_\D}(s,a)],$$
then pretraining approximately satisfies Eq.~\ref{eq:iql_D}, placing it in the \Superior or \Comparable regime; otherwise, it falls into the \Inferior regime.

In practice, we estimate $Q^{\pi_{\mathcal{D}}}(s,a)$ using the Monte Carlo return $G(s,a)$ from $\mathcal{D}$. We then perform a $t$-test to assess whether $\mathbb{E}_{(s,a) \sim \D}\!\left[\,\hat{Q}^{\pi_0}(s,a) - G(s,a)\,\right]$ is zero. Acceptance of the null corresponds to the \Comparable Q-regime; a significantly positive difference indicates the \Superior Q-regime, and a significantly negative difference indicates the \Inferior Q-regime. 

Finally, we evaluate how well this Q-based taxonomy aligns with the fine-tuning results. 
The corresponding confusion matrix is reported in Table~\ref{table:q-regime-confusion}.
It achieves 32 out of 63 correct predictions (\textbf{51\%}), 19 opposite mismatches (30\%), and 12 adjacent mismatches (19\%). Although this accuracy is substantially better than random guessing (33\%), it remains noticeably lower than the performance of our framework (\textbf{71\%}).

\begin{table}[htb]
\small
\centering
\caption{Confusion matrix of fine-tuning results using \textbf{Q-based taxonomy}. 
Green cells: correct predictions (32/63); red cells: opposite mismatches (19/63); gray cells: adjacent mismatches (12/63). Overall, Q-based taxonomy achieves 51\% correct predictions with 30\% opposite mismatches.}
\vspace{0.5em}
\label{table:q-regime-confusion}
\renewcommand{\arraystretch}{1.2}
\begin{tabular}{c|c|ccc}
\toprule
\multicolumn{2}{c|}{} & \multicolumn{3}{c}{\textbf{Pretraining Q-based Regime}} \\
\cmidrule(lr){3-5}
\multicolumn{2}{c|}{} & \Superior & \Comparable & \Inferior \\
\midrule
\multirow{3}{*}{\rotatebox{90}{\textbf{Fine-tune}}}
& $\pi_0$-centric $>$ $\mathcal{D}$-centric & \cellcolor{lightgreen}\textbf{23} & \cellcolor{lightgray}0 & \cellcolor{lightred}4 \\
& $\pi_0$-centric $\approx$ $\mathcal{D}$-centric & \cellcolor{lightgray}8 & \cellcolor{lightgreen}\textbf{0} & \cellcolor{lightgray}3 \\
& $\pi_0$-centric $<$ $\mathcal{D}$-centric & \cellcolor{lightred}15 & \cellcolor{lightgray}1 & \cellcolor{lightgreen}\textbf{9} \\
\bottomrule
\end{tabular}
\end{table}

\subsection{Behavior-Cloning-Based Taxonomy}

Instead of relying on the return of the behavior policy $J(\pi_\D)$, one can define a taxonomy based on the performance of a behavior-cloned policy $\pi_{\text{BC}}$ learned on the offline dataset $\D$. 

Similar to our taxonomy, we use a $t$-test with a margin of $\delta = 0.05$ to assess whether $J(\pi_{\text{BC}})$ and $J(\pi_{\mathcal{D}})$ differ significantly, thereby identifying the three corresponding regimes. The corresponding confusion matrix is shown in Table~\ref{table:bc-regime-confusion}. The BC-based taxonomy achieves 26 out of 63 correct predictions (\textbf{41\%}), 3 opposite mismatches (5\%), and 34 adjacent mismatches (54\%). While better than random guessing, this result remains noticeably lower than the performance of our framework (\textbf{71\%}).

\begin{table}[htb]
\small
\centering
\caption{Confusion matrix of fine-tuning results using \textbf{BC-based taxonomy}. 
Green cells: correct predictions (26/63); red cells: opposite mismatches (34/63); gray cells: adjacent mismatches (3/63). Overall, BC-based taxonomy achieve 41\% correct predictions with 5\% opposite mismatches.
}
\vspace{0.5em}
\label{table:bc-regime-confusion}
\renewcommand{\arraystretch}{1.2}
\begin{tabular}{c|c|ccc}
\toprule
\multicolumn{2}{c|}{} & \multicolumn{3}{c}{\textbf{BC-based Regime}} \\
\cmidrule(lr){3-5}
\multicolumn{2}{c|}{} & \Superior & \Comparable & \Inferior \\
\midrule
\multirow{3}{*}{\rotatebox{90}{\textbf{Fine-tune}}}
& $\pi_0$-centric $>$ $\mathcal{D}$-centric & \cellcolor{lightgreen}\textbf{17} & \cellcolor{lightgray}10 & \cellcolor{lightred}0 \\
& $\pi_0$-centric $\approx$ $\mathcal{D}$-centric & \cellcolor{lightgray}4 & \cellcolor{lightgreen}\textbf{6} & \cellcolor{lightgray}1 \\
& $\pi_0$-centric $<$ $\mathcal{D}$-centric & \cellcolor{lightred}3 & \cellcolor{lightgray}19 & \cellcolor{lightgreen}\textbf{3} \\
\bottomrule
\end{tabular}
\end{table}

\clearpage
\section{Extended Mechanistic Analysis}
\label{appendix:extended-mechanistic}
\begin{figure*}[h]
    \centering
    \includegraphics[width=0.85\linewidth]{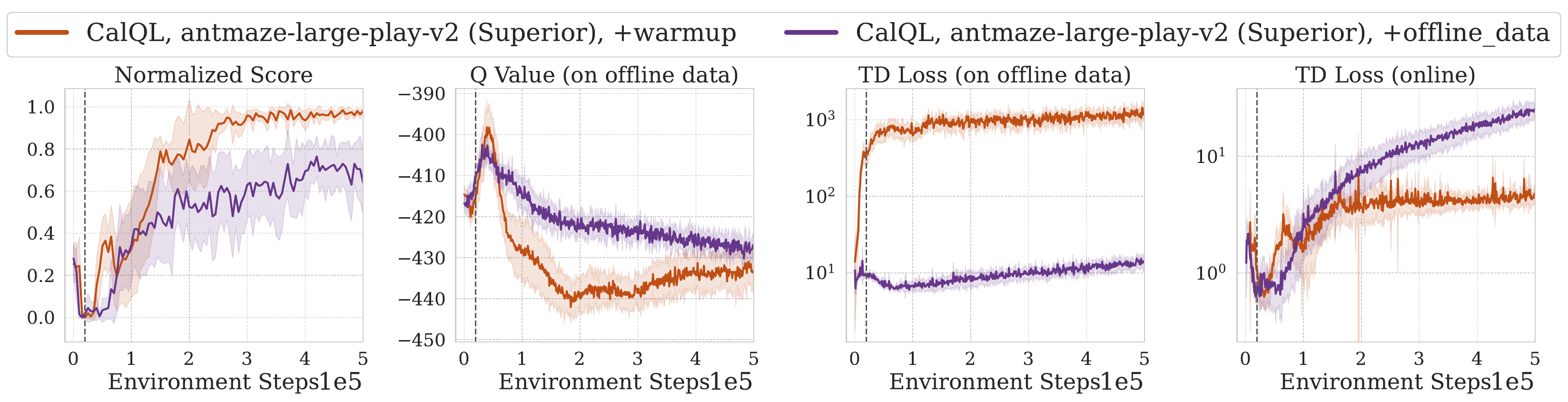}
\includegraphics[width=1.0\linewidth]{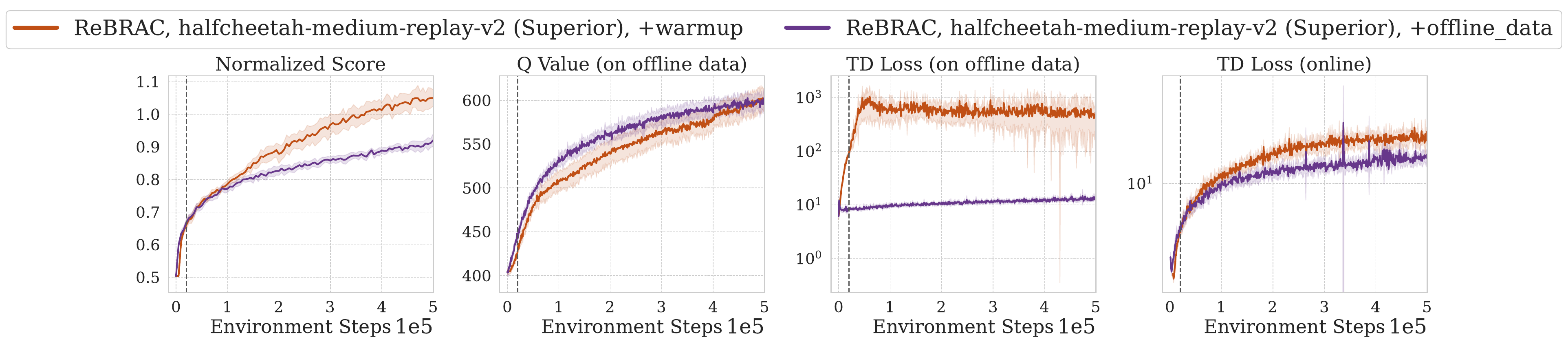}    
    \caption{Extended mechanistic analysis in the \Superior regime.}
    \label{fig:extended_superior}
\end{figure*}

\begin{figure*}[h]
    \centering
    \includegraphics[width=0.85\linewidth]{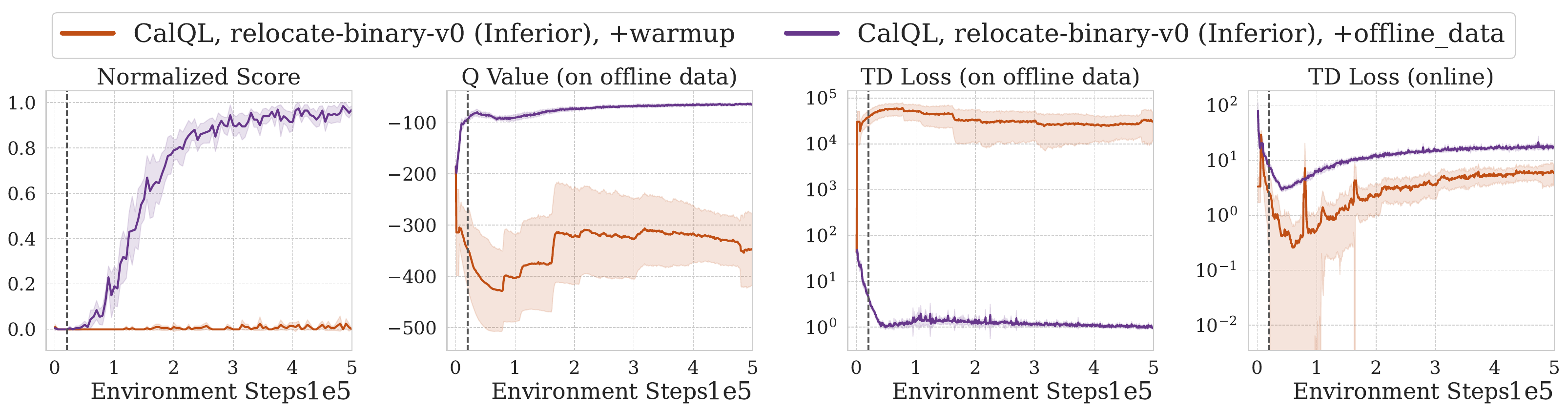}
    \includegraphics[width=0.85\linewidth]{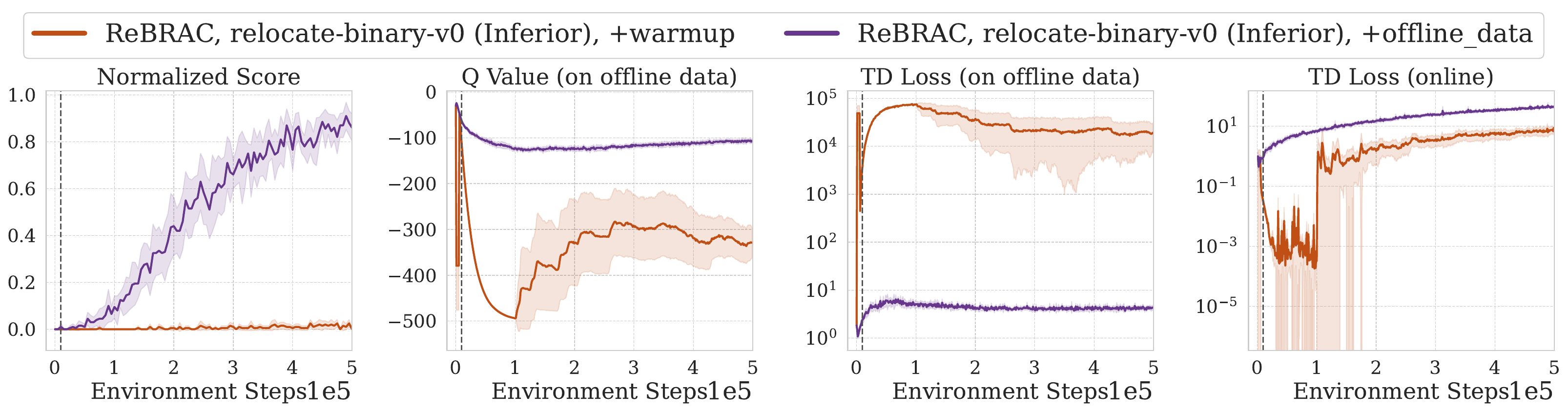}
    \caption{Extended mechanistic analysis in the \Inferior regime.}
    \label{fig:extended_inferior}
\end{figure*}

In this section, we extend our mechanistic analysis to a broader range of experimental settings. Figures~\ref{fig:extended_superior} and \ref{fig:extended_inferior} present the extended results for the \Superior and \Inferior regimes, respectively. Consistent with our main text discussion, these results illustrate that fine-tuning without explicit offline data anchoring leads to diverging Q-values and exploding TD errors on the offline dataset. While a similar pattern also occurs in the \Superior regime, it is significantly less severe.

\end{document}